\documentclass[10pt,twocolumn,letterpaper]{article}

\usepackage{cvpr}
\usepackage{times}
\usepackage{epsfig}
\usepackage{graphicx}
\usepackage{amsmath}
\usepackage{amssymb}

\usepackage{verbatim}
\usepackage{epstopdf}
\usepackage{helpers}

\usepackage{algorithm,algcompatible,amsmath}
\algnewcommand\INPUT{\item[\textbf{Input:}]}%
\algnewcommand\OUTPUT{\item[\textbf{Output:}]}%

\usepackage[pagebackref=true,breaklinks=true,letterpaper=true,colorlinks,bookmarks=false]{hyperref}

\cvprfinalcopy 

\def\cvprPaperID{1388} 

\ifcvprfinal\fi
\begin{document}

\title{Towards Building an RGBD-M Scanner}

\author{Zhe Wu\\
DJI\\
Shenzhen, China\\
{\tt\small wuzhe06@gmail.com}
\and
Sai-Kit Yeung\\
SUTD\\
Singapore\\
{\tt\small saikit@sutd.edu.sg}
\and
Ping Tan\\
Simon Fraser University\\
Vancouver, Canada\\
{\tt\small pingtan@sfu.ca}
}

\maketitle

\begin{abstract}
	We present a portable device to capture both shape and reflectance of an indoor scene. 
	Consisting of a Kinect, an IR camera and several IR LEDs, our device allows the user to acquire data
	in a similar way as he/she scans with a single Kinect. Scene geometry is reconstructed by
	KinectFusion. To estimate reflectance from incomplete and noisy observations, 3D vertices of the same 		
	material are identified by our material 
	segmentation propagation algorithm. Then BRDF observations at these vertices are merged into a more 
	complete and 
	accurate BRDF for the material. Effectiveness of our device is demonstrated by quality results on real-
	world scenes.
\end{abstract}

\section{Introduction}

Appearance capture involves simultaneous acquisition of both 3D shape and reflectance
of an object or a scene. 
Given such information, it is possible to produce photo-realistic
images for movies and games. Besides, it has seen wide application in both reverse engineering and cultural heritage preservation.
Because of its importance, various methods on this topic were proposed and some of them~\cite{tunwattanapong2013acquiring, zhou2013multi, holroyd2010coaxial} have achieved highly accurate results. 
However, these methods typically relied on giant and expensive hardware setups, and the data acquisition process required 
certain amount of expertise in this field. Lack of both has prohibited average users who wish to digitize objects or scenes
in everyday life from doing so. Another major issue of existing systems is that they are only targeted at acquisition of single movable objects of small scale(a few centimeters in diameter), which further prohibits their application in situations where a static and relatively large indoor scene is of interest to the user. Thus, a novel acquisition device which addresses these limitations is highly desirable for casual appearance capture.

Compared with appearance capture, acquisition of only 3D shape is a relatively easier task with the aid of Microsoft Kinect. Kinect is a portable and easy-to-use device which provides streams of depth images and RGB images of a scene.
By making use of the depth stream, \cite{izadi2011kinectfusion} achieved real-time 3D reconstruction. Per-vertex color information can be estimated by taking the RGB stream into account~\cite{zhou2014color}. Further efforts had been made by \cite{wu2014real, CPTK14cvpr, haque2014high, yu2013shading} to improve the quality of depth images or the reconstructed shape via photometric techniques using the RGB stream. However, none of them is capable of
revealing reflectance information, which is of great importance for a computer to better analyze and understand the scene.

In view of the above mentioned gaps in both fields, we present in this paper a novel device for appearance capture. Specifically, we equip ASUS Xtion Pro Live (a similar product to Kinect) with an additional infrared camera and several infrared LEDs, which are synchronized by a customized circuit. After acquiring data using this portable device, we first reconstruct the 3D scene from depth images by KinectFusion and then collect reflectance
information of each 3D vertex by projecting them onto the images of the IR camera under IR illumination. While BRDF observations for each 3D vertex are sparse and severely corrupted, we manage to group together vertices of the same material through our proposed algorithm called \textit{BRDF segmentation propagation}. By concatenating BRDF observations of vertices of the same material, we obtain a more complete and accurate estimation of BRDF for each material in the scene.
Our system can be regarded as an RGBD-M(aterial) scanner due to its capability of depth sensing and reflectance sensing.

In summary, the contributions of this paper are twofold.

1. We prototype a novel appearance acquisition device featuring portability and ease of use. Moreover, this device is able
to capture the shape and reflectance of a relatively large indoor scene instead a single object;

2. We present a novel method for automatic classification of 3D vertices based solely on their sparse and corrupted BRDF observations.

\section{Related Work}
\mycomment{
The system presented in this paper is aimed at simultaneous acquisition of
shape and reflectance with the aid of customized hardware. It is
related to the fields as reviewed below.
}

\noindent
\textbf{Appearance Capture}
Appearance capture methods can be classified into
two categories based on input data. Methods in the first category~\cite{lensch2003image, sato1997object} rely on a range sensor to first reconstruct a precise 3D shape.
By collecting observations from an additional RGB camera and fitting parametric BRDFs, it is possible
to reconstruct the reflectance information at each surface point. Methods of the second category involve only RGB cameras to reconstruct both 3D shapes and BRDFs. \cite{hernandez2008multiview, goldman2010shape, alldrin2008photometric, zhou2013multi} made various assumptions on BRDFs to produce high-quality results while \cite{holroyd2010coaxial, tunwattanapong2013acquiring} achieved the same goal by smartly designing the acquisition system: using either a coaxial optical scanner or spherical harmonic illumination. The major advantage of such systems is accuracy. However, accuracy comes at the expense of giant and complicated hardware setups in a controlled environment. Besides, these systems worked in the visible spectrum and required a darkroom environment when capturing data. The third issue for such systems is that they work only on small-scale objects which can be placed on a turntable. This issue prohibits them from being used for acquiring the appearance of an indoor scene.

\noindent
\textbf{Dense Scene Reconstruction}
3D reconstruction has been one of the key problems in computer vision.
By using solely RGB images, multi-view stereo methods such as~\cite{furukawa2010accurate} 
and structure-from-motion(SfM) methods like~\cite{newcombe2011dtam} could produce plausible 3D shape models. 
The emergence of Microsoft Kinect, which has a consumer-level range sensor, enables novel approaches~\cite{ izadi2011kinectfusion, zhou2013dense} to dense 3D 
reconstruction. Color information can also be estimated with an additional RGB camera~\cite{zhou2014color}.

3D reconstruction by Kinect cannot preserve fine surface details. This problem was tackled with the help of the RGB camera on Kinect. \cite{yu2013shading, haque2014high, CPTK14cvpr, wu2014real} either refined each depth image or refined the final triangular mesh by photometric stereo approaches.
One recent work~\cite{zollhofershading} refined geometry directly on the volumetric representation of the shape instead of the explicit mesh representation.
While these methods achieved plausible results in revealing detailed shape information, BRDFs, which play an important role in relighting and scene understanding, remain missing.

\mycomment{
\noindent
\textbf{Computational Sensing}
Traditional RGB cameras provide users with a color image. In recent years, there has been a trend of modifying the RGB 
camera or augmenting it with additional hardware to enhance its imaging capability. \cite{raskar2004non} made a non-photorealistic camera out of an ordinary RGB camera by augmenting it with synchronized flashes. \cite{liu2014discriminative}
enabled an RGB camera to differentiate materials by adding active illumination sources controlled by a computer. 
A recent work \cite{tang2014high} extended traditional RGB mosaic to allow IR light to be received by the camera sensor, 
and the additional IR channel enabled high resolution photography.
Kinect can also be viewed as a new computational sensing device due to its capability of depth sensing.
Compared with Kinect, our device takes one step further to recover not only shape, but also BRDFs of a scene. Because of this capability, our device can be seen as an RGBD-M(aterial) scanner in a general sense.
}

The work closest to ours is \cite{wu2015appfusion}, which made use of a single Kinect for appearance capture.
While both systems feature portability and ease of use, ours is different from that of \cite{wu2015appfusion} in following ways.

\noindent
1. \cite{wu2015appfusion} worked on a single object because of the requirement of environmental lighting while our device works on scenes, such as a room's corner, thanks to its active illumination in the IR spectrum;

\noindent
2. \cite{wu2015appfusion} assumed a parametric BRDF model while we use a bivariate BRDF model which is deduced from reflectance symmetries and is represented as a 2D table. Compared with a parametric model, the bivariate model is applicable to a wider range of real-world materials.

\noindent
3. \cite{wu2015appfusion} required illumination calibration whenever the visible illumination changes, which is done by placing a mirror sphere into the scene, while our device does not require such calibration;

\noindent
4. \cite{wu2015appfusion} required a user-specified number of materials, which is hard for average users to decide, while our system does not require additional input from the user.

\section{Hardware Description}
In this section, our customized device is first introduced. Then we briefly mention various calibrations involved 
before using the device. Finally, the data acquisition process is 
presented to illustrate the device's ease of use.

\subsection{Device Setup}

\begin{figure}
\begin{center}
\addtolength{\tabcolsep}{-4pt}  
\begin{tabular}{cc}
  \raisebox{-0.5\height}{\includegraphics[width=0.5\linewidth]{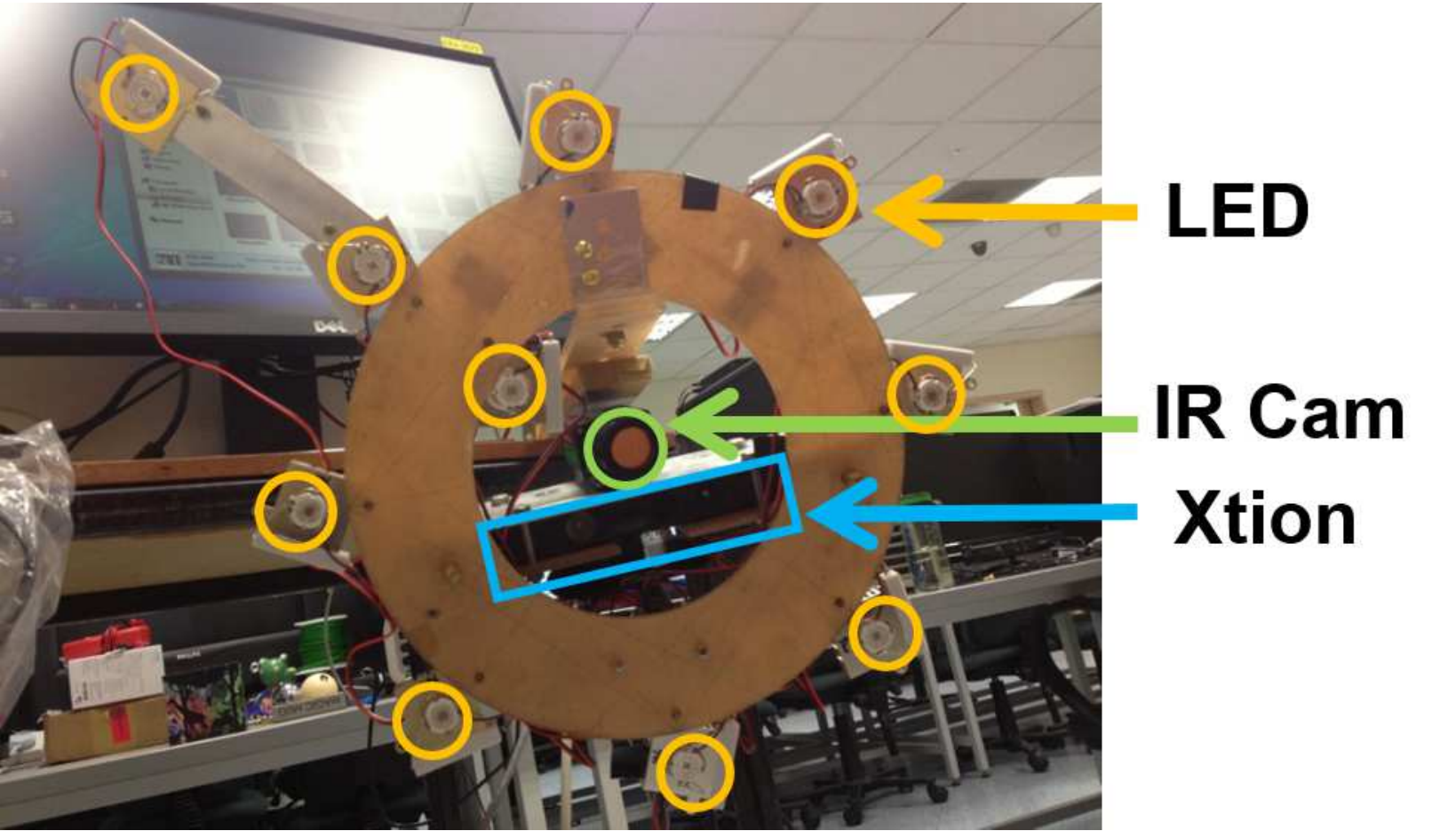}} & \raisebox{-0.5\height}{\includegraphics[width=0.5\linewidth]{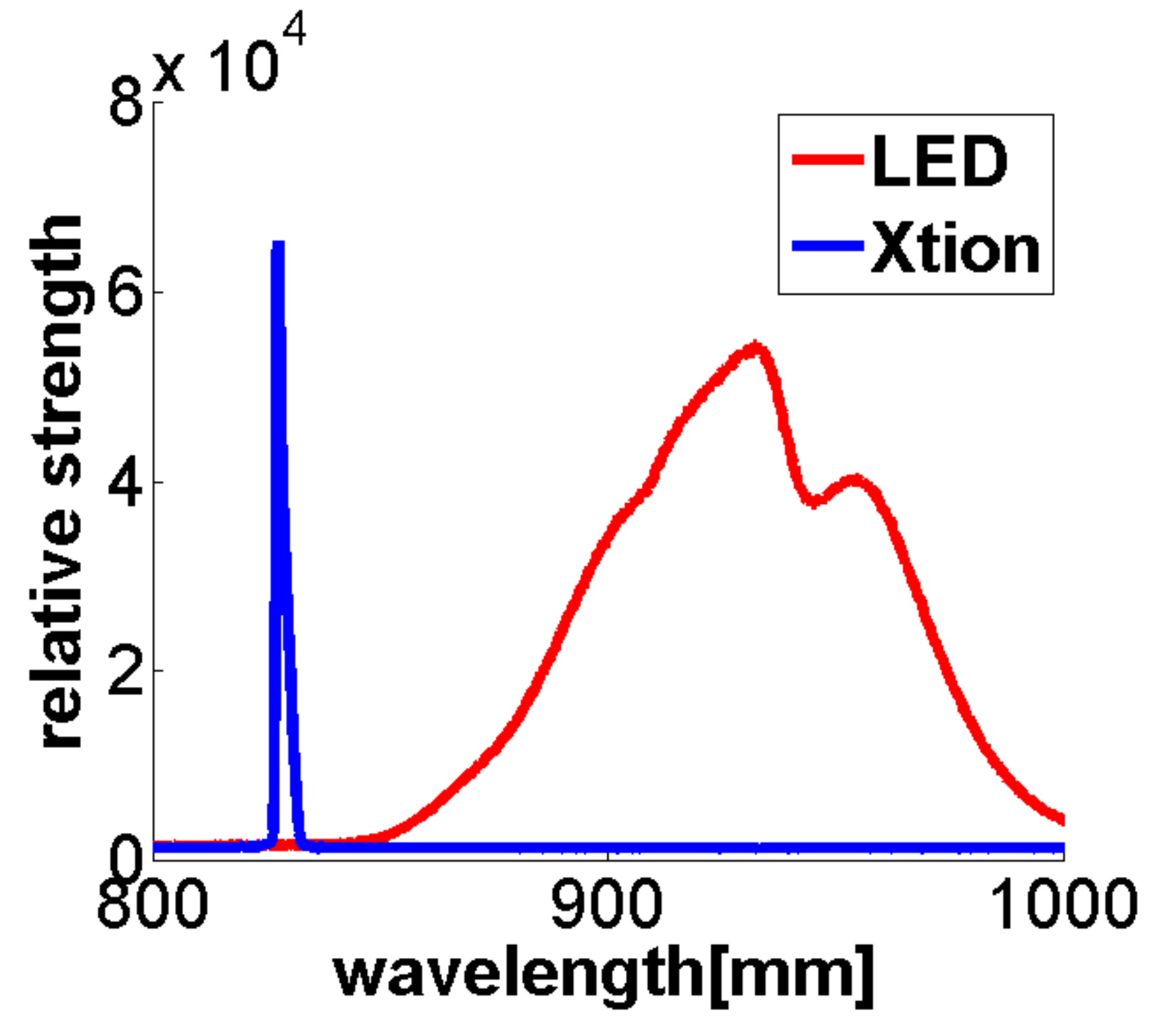}} \\
(a) & (b)
\end{tabular}
\addtolength{\tabcolsep}{4pt}  
\end{center}
\caption{(a) shows our prototype device; (b) shows the spectrum of Xtion and IR LEDs}
\label{device_spectrum}
\vspace{-6mm}
\end{figure}

\begin{figure*}
\centering
\includegraphics[width=0.9\linewidth]{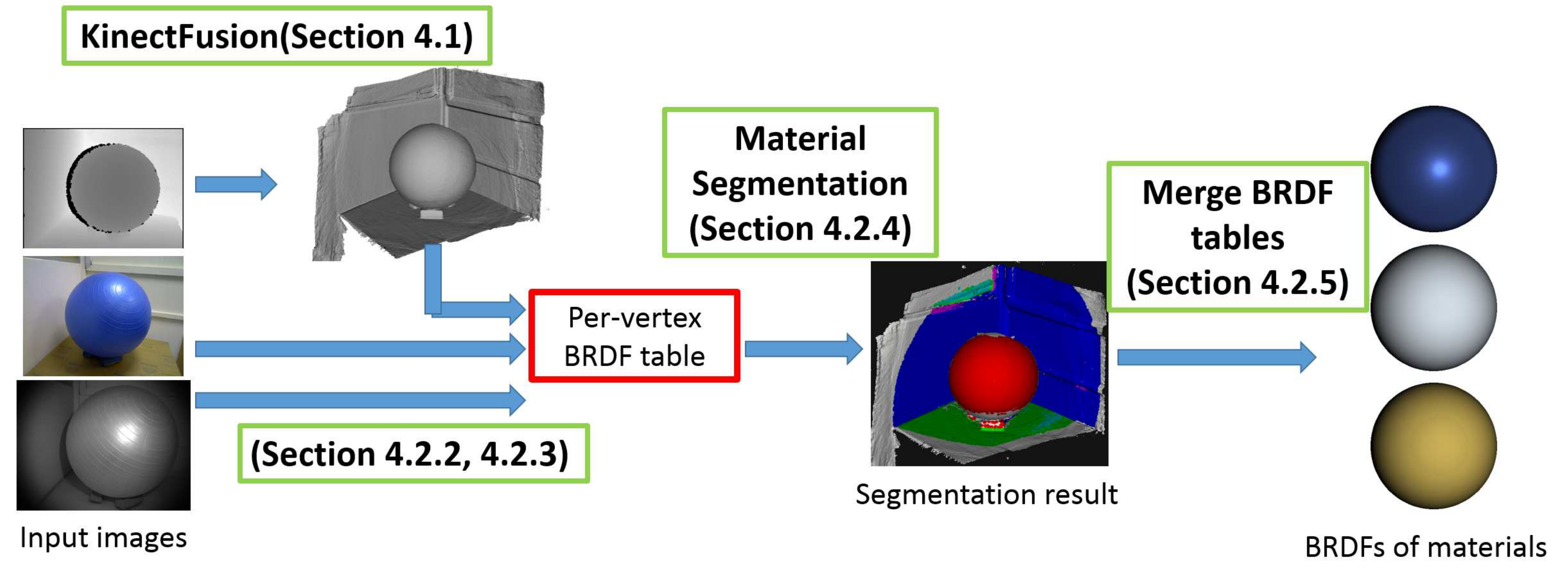}
\caption{System pipeline}
\label{pipeline}
\vspace{-5mm}
\end{figure*}

As shown in \figref{device_spectrum}(a), our customized device involves three key parts. The first one is ASUS Xtion PRO LIVE, or Xtion for short. Similar to Microsoft Kinect, Xtion streams depth and RGB images, 
whose resolutions are both $480 \times 640$, at 30 fps. 

\mycomment{However, Xtion is preferred in our setup because of its smaller form factor and the fact that it uses a single USB port to transmit both data and power, which is simpler than Kinect that requires a dedicated cable for power supply.}

The second part is an infrared(IR) camera which is sensitive in the near IR spectrum($750 \scriptsize{\sim} 1000nm$). The IR camera and Xtion are fixed on opposite sides of a thick Aluminium plate so that the baseline between them is minimized. The resolution of the IR camera is $518 \times 694$ and the lens was chosen so that the IR camera and Xtion have similar FOVs.

The third part is a set of 10 IR LEDs. These LEDs are synchronized with the IR camera by a customized control circuit in such a way that LEDs are switched on and off sequentially while the IR camera acquires an IR image when a single LED is switched on. 
\mycomment{To reduce the effects of underexposure and overexposure, for each LED, we capture two images under different exposures ($10ms$ and $30ms$).} 
The frame rate for the IR camera is around 20 fps.
LEDs are fixed on a circular plastic plate, with eight of them evenly distributed on a circle with a radius of $16cm$, the center of which is the IR camera, and the rest two found at $8cm$ and $28cm$ away from the IR camera respectively. Such kind of lighting design is to facilitate robust BRDF estimation.

\mycomment{
Since Xtion's depth camera also works in the IR spectrum, we have to carefully avoid interference between Xtion and the IR LEDs. We measured the spectrum of Xtion's depth camera by a spectrometer, which suggests a peak at around $835nm$. Thus, we chose IR LEDs with a peak at around $940nm$ in the spectrum as shown in \figref{device_spectrum}(b). Clearly, Xtion's depth camera will be not affected by the additional IR LEDs. However, the IR camera collects photons from the visible spectrum to near IR spectrum. Thus, we placed a \textit{bandpass filter} in front of the IR camera so that it only receives light with a wavelength at around $940nm$.
}

Since Xtion's depth camera also works in the IR spectrum, we carefully chose the working spectrum of IR LEDs to minimize interference between Xtion and IR LEDs, as shown in \figref{device_spectrum}(b). Besides, a \textit{bandpass filter} was placed in front of the IR camera so that it only receives light from IR LEDs.

Both Xtion and the IR camera are connected to a desktop computer and controlled by a data capture program.

\subsection{Device Calibration}
Our device was calibrated in the following aspects.

1. We assume perspective camera model for the IR camera, Xtion's depth camera and RGB camera, and we calibrated
their intrinsic matrices. We also calibrated the relative poses between any pair out of the three;

2. An LED is considered a point light source and its relative position to the IR camera was calibrated;

3. Vignette effect found in IR images was calibrated;

4. Relative brightness of LEDs were calibrated.

Another issue is temporal synchronization between Xtion and the IR camera. We simply record a timestamp for every image received by the computer. These timestamps will be used in \secref{per_vertex_ir_brdf}.

\subsection{Data Acquisition}

Our device is intended to be used in a similar way as we use a single Kinect: the user holds the device and points it to the scene of interest. Moving around the scene, our device generates three image streams: depth stream, RGB stream and IR stream. These images are transmitted back to a computer and will later be used to reconstruct the appearance model of the scene.

\section{Appearance Capture}

Given acquired depth, RGB and IR images, our system estimates both shape and reflectance of the scene.
\figref{pipeline} shows the processing pipeline of our system. In this section, we detail each step.

\subsection{Geometry Reconstruction}
We first obtain the 3D shape of the scene by KinectFusion on the sequence of depth images.
The output is a triangular mesh representing the shape, together with a camera pose for every depth image.

\subsection{Reflectance Estimation}
For a surface point in the scene, we try to obtain its reflectance information, which is described by a 
BRDF(bidirectional reflectance distribution function) $f(\omega_{in}, \omega_{out}, \lambda)$, where $\omega_{in}$
is the incoming lighting direction, $\omega_{out}$ is the outgoing light direction, and $\lambda$ is the wavelength of the light.

\subsubsection{Reflectance Models}

While one material may not exhibit exactly the same reflectance characteristic in different ranges of the light spectrum, the difference is negligible for dielectrics in the range of visible and near-infrared spectrum according to~\cite{wu2015appfusion}. Thus, we simplify the BRDF model as $f_{\mathrm{IR}}(\omega_{in}, \omega_{out}) (R,G,B)^T$, where $f_\mathrm{{IR}}$ is the BRDF estimated in the IR spectrum, $(R,G,B)^T$ is the \textit{normalized} color vector estimated in the visible spectrum.

A general $f_{\mathrm{IR}}(\omega_{in}, \omega_{out})$ is a 4D function and requires much effort to be fully captured. Instead of adopting oversimplified parametric models, we assume isotropy and half-vector symmetry of BRDFs. Then the 4D function can be reduced to a bivariate BRDF
\begin{equation}
f_{\mathrm{IR}}(\omega_{in}, \omega_{out}) = f_{\mathrm{IR}}(\theta_h, \theta_d)
\end{equation}
where $\theta_h$ or $\theta_d$ are the half angle and difference angle defined in~\cite{rusinkiewicz1998new}.
It is worth noting that isotropy and/or half-vector symmetry have been used extensively in previous work on photometric stereo~\cite{alldrin2008photometric, romeiro2008passive, wu2013calibrating}.

We further assume that the scene contains only a few different BRDFs. This assumption is a good approximation of an indoor scene consisting of artificial furniture and articles~\cite{yu1999inverse}, which is exactly the target scene of our system.

Note that we do not adopt the dichromatic BRDF model because it increases the complexity of our BRDF model, which makes it prone to large noises observed in our data.

\noindent
\textbf{BRDF Representation}
We represent a BRDF as a 2D lookup table. $\theta_h \in [0^o, 90^o]$ is discretized into 45 bins of equal width while $\theta_d \in [0^o,90^o]$ is discretized into 48 bins. Thus, the BRDF representation is a $45 \times 48$ table with each entry containing a 3-vector $(Rf_{\mathrm{IR}}, Gf_{\mathrm{IR}}, Bf_{\mathrm{IR}})^T$. We note an entry of the table as a \textit{BRDF cell}.
Each vertex is associated with such a BRDF table.

\subsubsection{Per-Vertex Color}
\label{sec:per_vert_color}
By projecting a 3D vertex onto an RGB image, the RGB values at the projected 2D pixel give an estimation of the 3D vertex's color.

\mycomment{
Given the pre-calibrated relative pose $T_{\mathrm{d \rightarrow RGB}}$ between the depth camera and the RGB camera, 
we can easily calculate the camera pose of every RGB image as follows
\begin{equation}
P_{\mathrm{RGB}, t} = T_{\mathrm{d \rightarrow RGB}} P_{\mathrm{Depth}, t}
\end{equation}
where $P_{\mathrm{RGB}, t}$ and $P_{\mathrm{Depth}, t}$ are the poses of the RGB camera and depth camera for a pair of RGB and depth images captured with the same timestamp $t$.
With the aid of the RGB camera's pose, a vertex can be projected into an RGB image, where the corresponding pixel's RGB values give an estimation of the vertex color.
}

After collecting a set of RGB values from a fraction of all RGB images, a 3D vertex's color is calculated
as the median of these RGB values on a per-channel basis. Note that we discard saturated RGB values and those 
estimates when viewing at a grazing angle($\omega_{out} > 60^o$).
The normalized per-vertex color $(R,G,B)^T$ is the result of normalization of the above estimated color.

\subsubsection{Per-Vertex IR BRDF}
\label{per_vertex_ir_brdf}
The BRDF of a vertex in the IR spectrum $f_{\mathrm{IR}}$ is estimated from IR images.
For a vertex $X$ on a surface of the scene, we can estimate one of its BRDF values $f_{\mathrm{IR}}(\theta_h, \theta_d)$ from an IR image $I_t$ acquired at time $t$ based on the 
following image formation model
\begin{equation}
\label{image_formation}
I_{t}(x) = \mathrm{vig}(x) \mathrm{vis}_{t}(X) f_{\mathrm{IR}}(\theta_h, \theta_d) (\mathbf{n}^T \mathbf{l_t}) \frac{L_t}{d_t^2},
\end{equation}
where $\mathbf{n}$ is the vertex $X$'s normal, $\mathbf{l_t}$ is the light direction for $X$ at time $t$, $L_t$ is the  LED's relative brightness, $d_t$ is the distance between the LED and $X$, and $x$ is the image coordinate when projecting $X$ onto the IR image.
The function $\mathrm{vig}(x)$ models vignette effect while $\mathrm{vis}_t(X) \in \{0,1\}$ encodes both visibility and shadow information.

It is worth noting that since we adopt the perspective camera model and point light source model, both $\theta_h$ and $\theta_d$ 
vary as our device moves. Thus, we are able to collect BRDF values at different $(\theta_h,\theta_d)$.

To facilitate the estimation, we need to first know the IR camera's pose, which, unfortunately, cannot simply be calculated from any single camera pose of the depth camera due to lack of frame-to-frame correspondence.
To deal with this, the IR camera's pose is interpolated from two temporally neighboring depth camera's poses, given known timestamps. Note that the rotational component of a camera pose can be interpolated via spherical linear interpolation of quaternion \cite{dam1998quaternions}.

\mycomment{
 in a similar way as we estimate an RGB image's pose
\begin{equation}
P_{\mathrm{IR},t} = T_{\mathrm{d} \rightarrow \mathrm{IR}} P_{\mathrm{Depth}, t}
\end{equation}
where $T_{\mathrm{d} \rightarrow \mathrm{IR}}$ is the pre-calibrated relative pose between the IR camera and the depth camera.

However, since the IR stream and depth stream do not have a frame-to-frame correspondence, $P_{\mathrm{Depth}, t}$ 
does not exist and can only be obtained through interpolation of two neighboring camera poses 
$P_{\mathrm{Depth}, t_0}$ and $P_{\mathrm{Depth}, t_1}$, where $t_0 < t < t_1$.

The interpolation of a camera pose, which is composed of a translational component and a rotational component, 
involves interpolation of both separately. The interpolation of translational vectors is simple linear interpolation
while the rotational interpolation is achieved through spherical linear interpolation of quaternion, which is a
representation of rotation. Details of spherical interpolation of quaternion can be found in~\cite{dam1998quaternions}.
}

After obtaining the camera pose of an IR image, the position of the IR image's corresponding LED can be easily obtained since all LEDs' positions in the IR camera's coordinate system are pre-calibrated. Thus, the IR BRDF value can be calculate based on \equref{image_formation}.

Given a vertex's normalized color $(R,G,B)$ from \secref{sec:per_vert_color} and an IR BRDF value $f_{\mathrm{IR}}(\theta_h,\theta_d)$ from this section, we can calculate its full BRDF vector at $(\theta_h,\theta_d)$ in the visible spectrum as $(Rf_{\mathrm{IR}}, Gf_{\mathrm{IR}}, Bf_{\mathrm{IR}})$. Multiple BRDF vectors at the same $(
\theta_h,\theta_d)$ are averaged before being placed into the vertex's associated BRDF table.

\subsubsection{Material Segmentation}
\label{set:material_segmentation}

\noindent
\textbf{Motivation}
Due to limited variation in the viewing/lighting directions, every single vertex has only
a sparse set of non-empty BRDF cells in its associated BRDF table.
Besides, unlike traditional BRDF acquisition systems \cite{wang2008modeling, aittala2013practical} where almost perfect calibration and registration are available, the BRDF values obtained using our device are \textit{severely corrupted} due to less accurate camera poses, imperfect shape-image registration and noisy surface normals. Thus, if we could identify 3D vertices of the same material, their BRDF tables can be merged into
a more complete and accurate BRDF estimation for that material.

If we consider the BRDF table as a feature vector for each vertex, this material segmentation problem can simply be cast as a spectral clustering problem by defining an affinity matrix for all vertices based on their BRDF tables. However, the affinity matrix is difficult to define because feature vectors have many different missing entries and the rest noisy entries are quite unreliable.

Another straightforward approach is to cluster vertices based on their colors. Mathematically, this amounts to feature dimension reduction, where the RGB vector is the empirical reduction of a whole BRDF table. Obviously, the RGB vector is not guaranteed to be the `optimal' reduction for clustering. In fact, different materials might share the same color. Even the same material might exhibit different colors when the lighting/viewing configuration changes(See the highlight on the gymball of \figref{gymball_results}(a)).

\mycomment{
Due to limited variation in the viewing/lighting directions, every single vertex has only
a sparse set of non-empty BRDF cells in its associated BRDF table.
If we could identify vertices of the same material, their BRDF tables would complement
each other and produce a more complete BRDF estimation.
}

\noindent
\textbf{Algorithm} In view of this, we propose a novel method for clustering vertices by utilizing
all the incomplete and noisy BRDF tables. We notice that while clustering all vertices at once is challenging, partial clustering within a single BRDF cell is relatively easy and more reliable even with large noise.
The basic idea of our algorithm is to segment vertices into different material groups based on their samples in the same BRDF cell, and then propagate this segmentation information into other BRDF cells to facilitate further segmentation of other vertices.
 
While there is usually an undefined number of materials in a real scene, we detail our method for the 2-material segmentation case here to illustrate our idea. The algorithm is presented in \algoref{segmentation_algo}, with each step detailed below.

\mycomment{
Multi-material segmentation shares the same idea and its implementation is omitted here for brevity. It is worth noting that our method does not require the 
user specify the number of materials in the scene.
}
\begin{algorithm}
    \caption{Two-material segmentation propagation}
  \begin{algorithmic}[1]
    \INPUT BRDF tables ${T_1, T_2, ...}$ of all vertices
    \OUTPUT Every vertex's material label
    \STATE Set up $T$ with initial clustering within each of its cells
    \STATE Select the most separable cell $\mathrm{Cell}_k$
    \STATE Separate $\mathrm{Cell}_k$ to initialize material groups $\mathrm{Mat}_1$ and $\mathrm{Mat}_2$
    \WHILE{ There are cells to be separated }
		\STATE Select currently most separable cell $\mathrm{Cell}_l$
		\STATE Separate $\mathrm{Cell}_l$ to update $\mathrm{Mat}_1$ and $\mathrm{Mat}_2$
    \ENDWHILE
  \end{algorithmic}
  \label{segmentation_algo}
\end{algorithm}

\begin{figure}
\begin{center}
\begin{tabular}{cc}
  \raisebox{-0.5\height}{\includegraphics[width=0.4\linewidth]{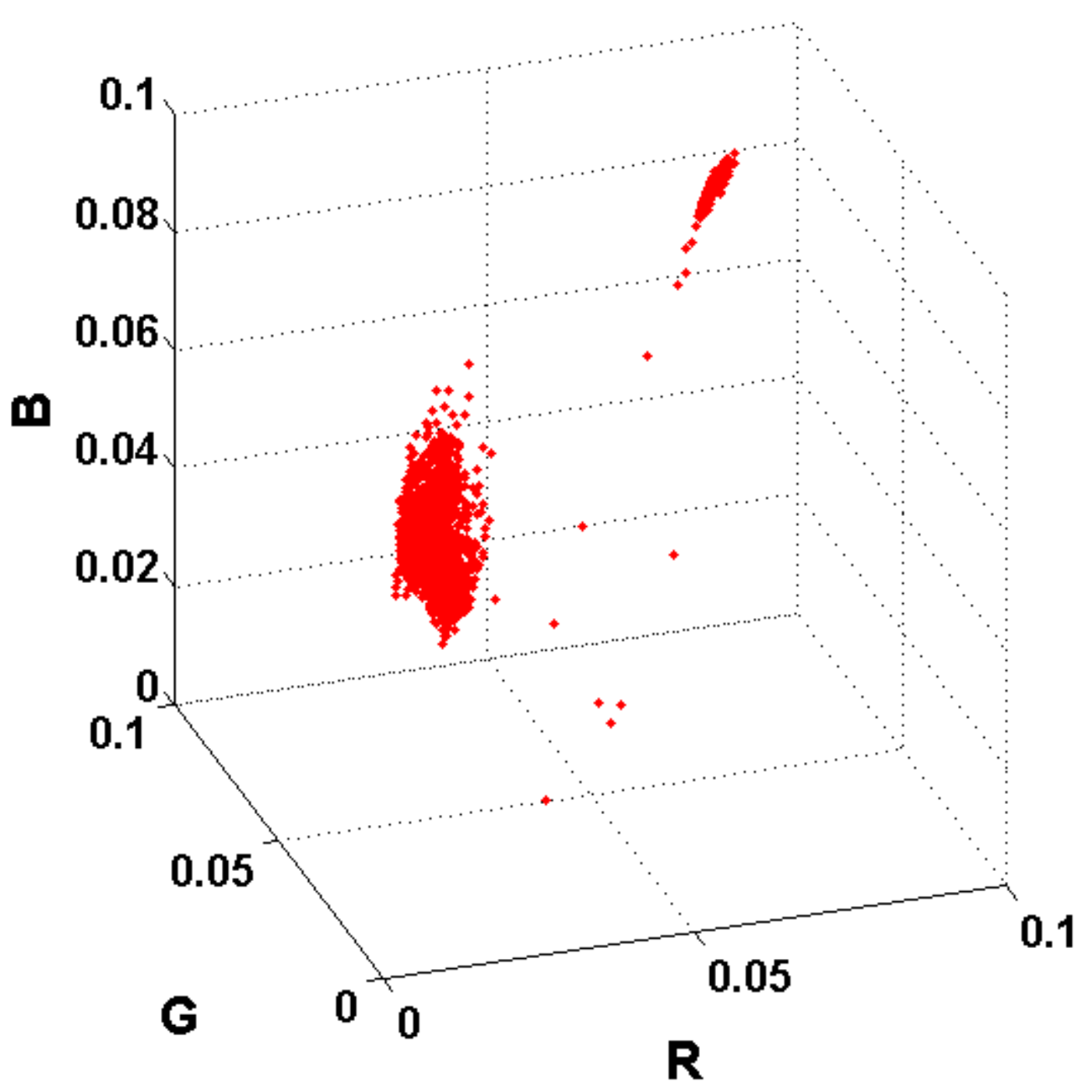}} & 
  \raisebox{-0.5\height}{\includegraphics[width=0.4\linewidth]{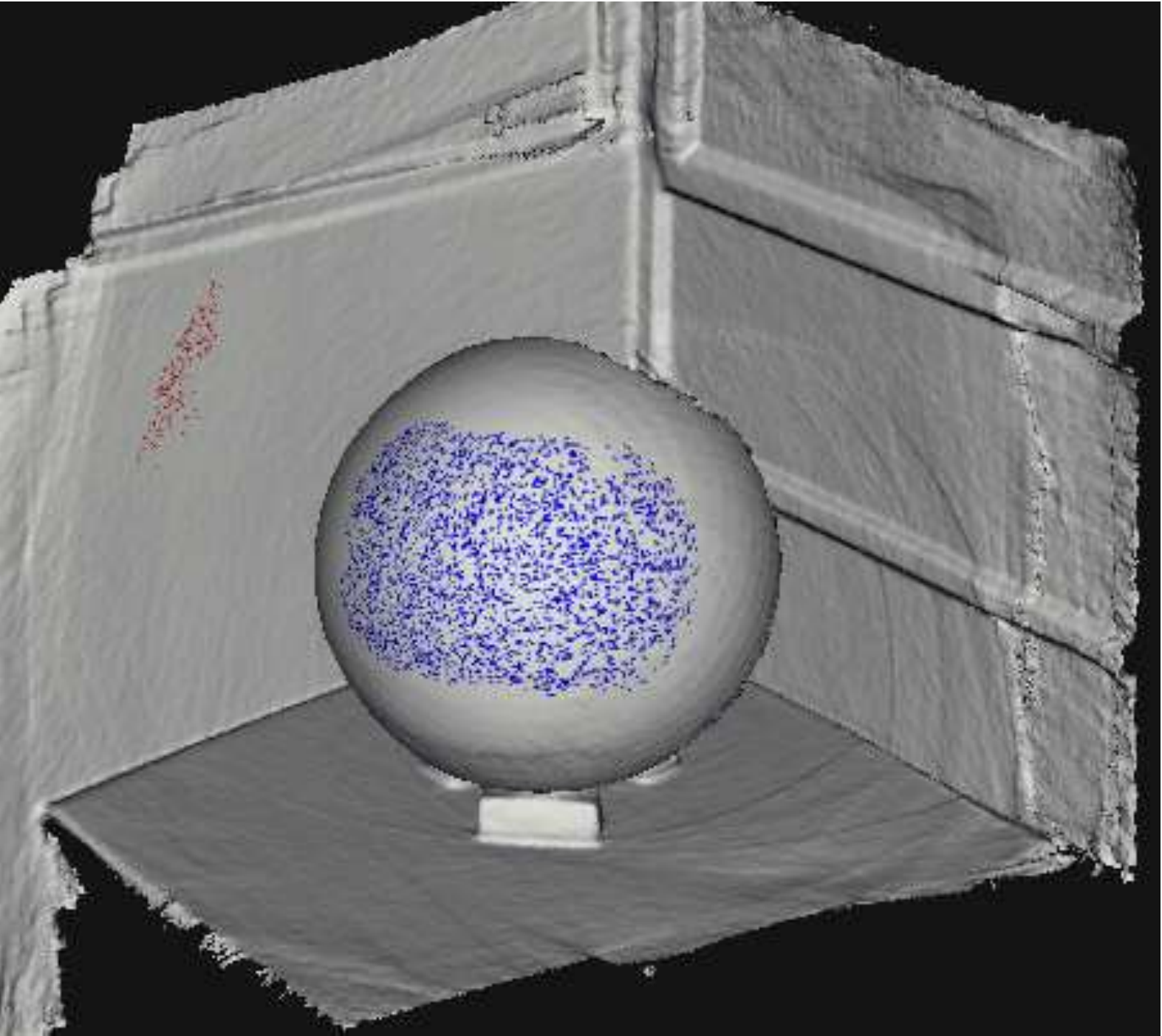}} \\
  (a) & (b) \\
  \raisebox{-0.5\height}{\includegraphics[width=0.4\linewidth]{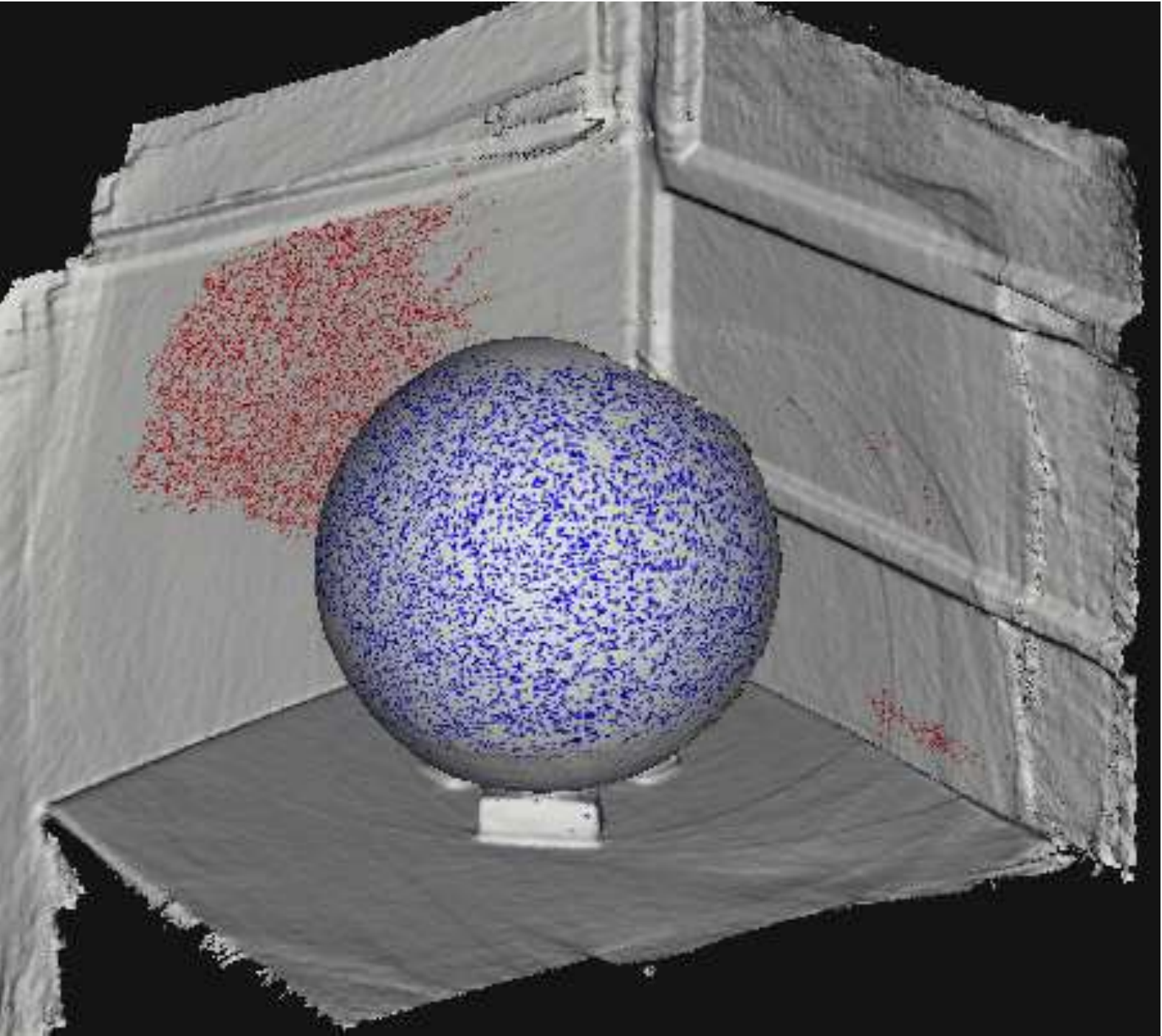}} &
  \raisebox{-0.5\height}{\includegraphics[width=0.4\linewidth]{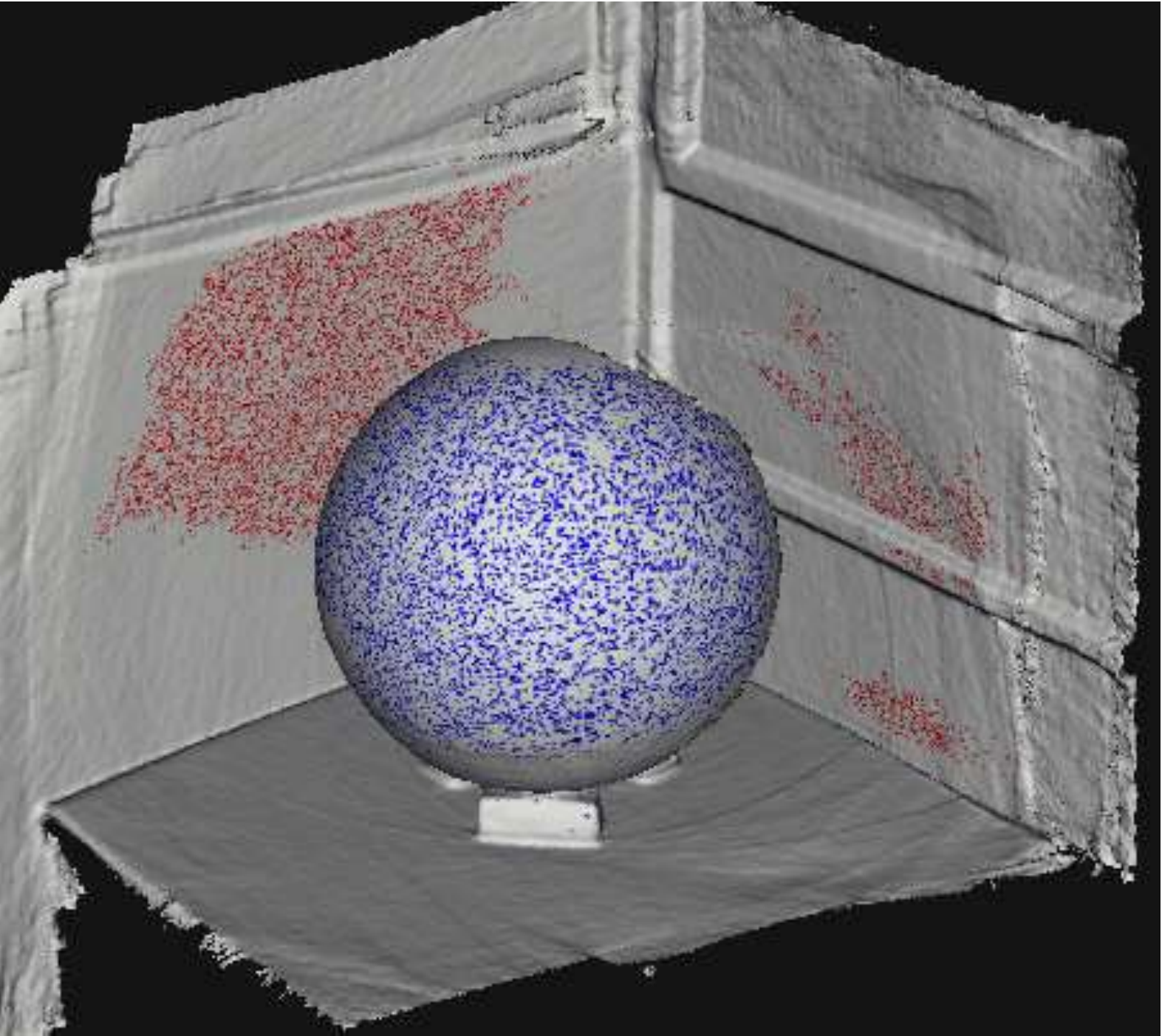}} \\
  (c) & (d) \\
  \raisebox{-0.5\height}{\includegraphics[width=0.4\linewidth]{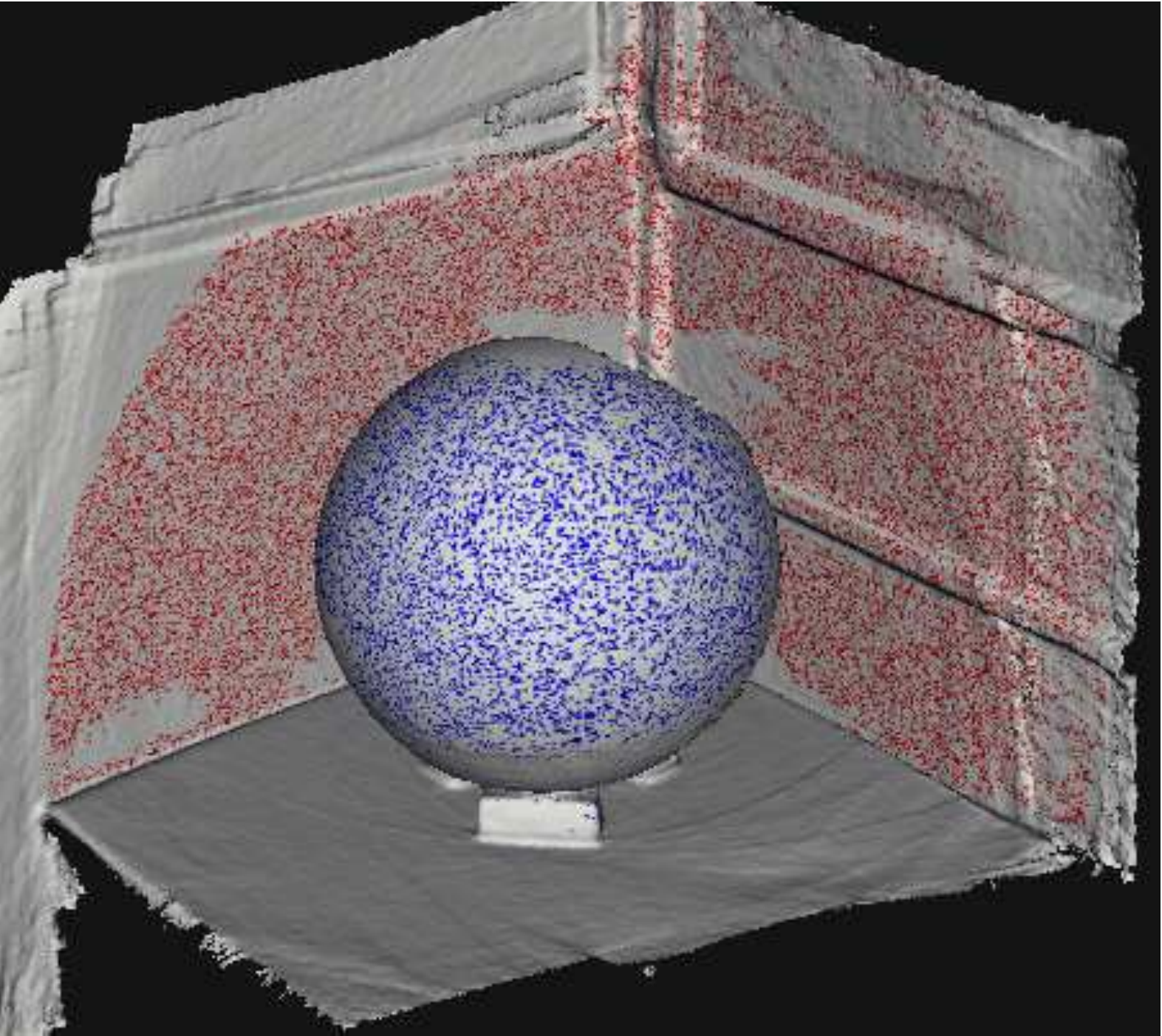}} &
  \raisebox{-0.5\height}{\includegraphics[width=0.4\linewidth]{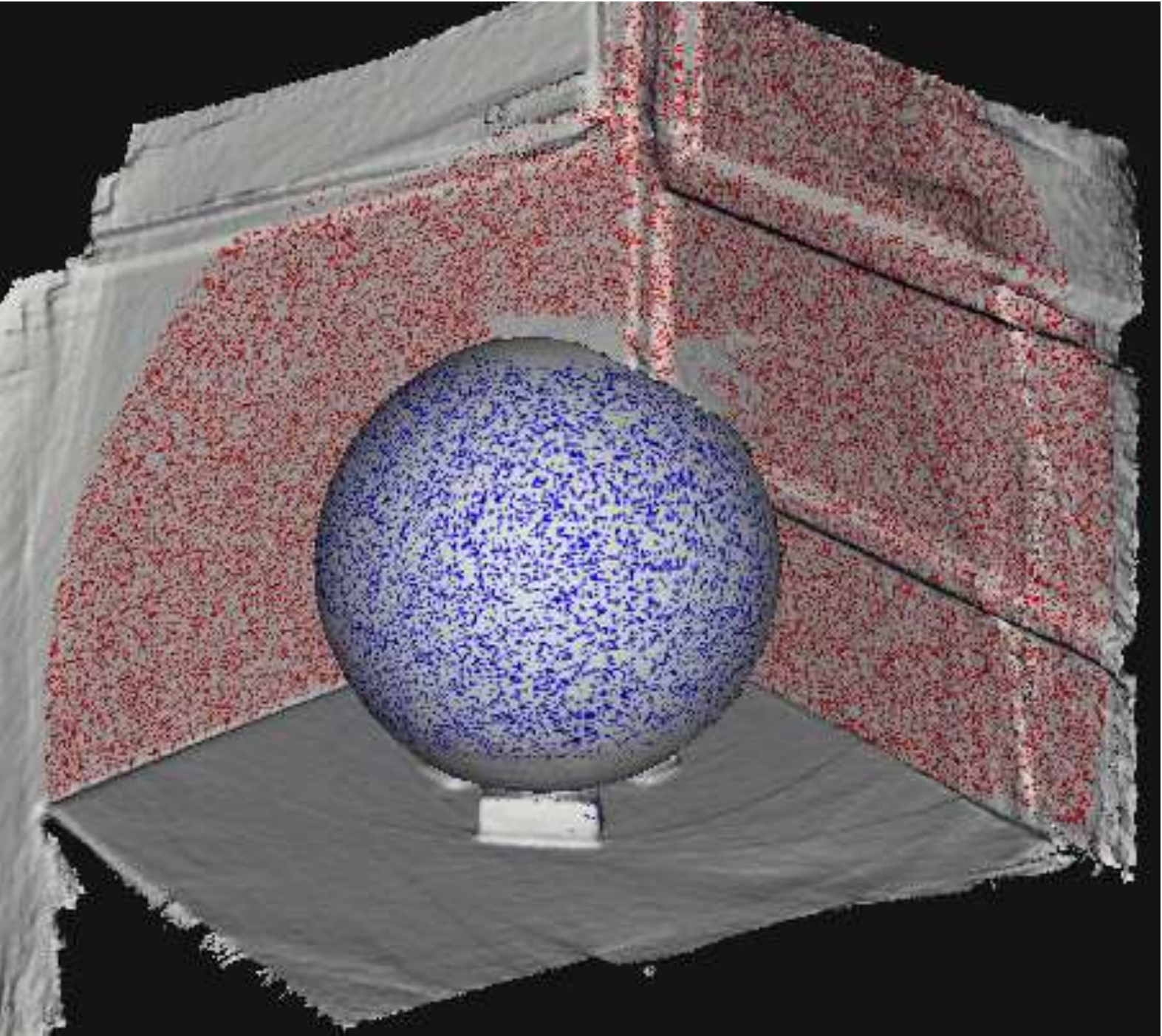}} \\
  (e) & (f)
\end{tabular}
\end{center}
\caption{Illustration of segmentation propagation. Note that while the scene involves over 2 materials, it can still
be used for demonstration because the initially segmented BRDF cell contains only two clusters}
\label{mat_segmentation}
\vspace{-5mm}
\end{figure}

\noindent
\textbf{Line 1}
We set up a new BRDF table $T$. A cell at $(\theta_h, \theta_d)$ of $T$ is associated with a list of BRDF samples collected from every vertex's corresponding BRDF cell if it is non-empty. \figref{mat_segmentation}(a) is a visualization of these samples at a cell in the RGB space. Note that only a random set of $100,000$ vertices of the mesh are used here for efficiency.

Ideally, BRDF samples in a cell of $T$ are merely repetitions of a few isolated points in the RGB space, corresponding to different materials. Due to various noises, however, these samples form a few clusters. We run the meanshift algorithm to automatically cluster these samples. The bandwidth of the meanshift algorithm is empirically set as $\frac{1}{2}\sqrt{V_{R}+V_{B}+V_{G}}$, where $V_R$, $V_G$ and $V_B$ are variances of samples along three dimensions respectively.
Those clusters with less than $5\%$ of all samples in the cell are discarded.

Note that in the 2-material case, there are no more than 2 clusters for each BRDF cell.

\begin{figure}
\begin{center}
\begin{tabular}{c}
  \raisebox{-0.5\height}{\includegraphics[width=\linewidth]{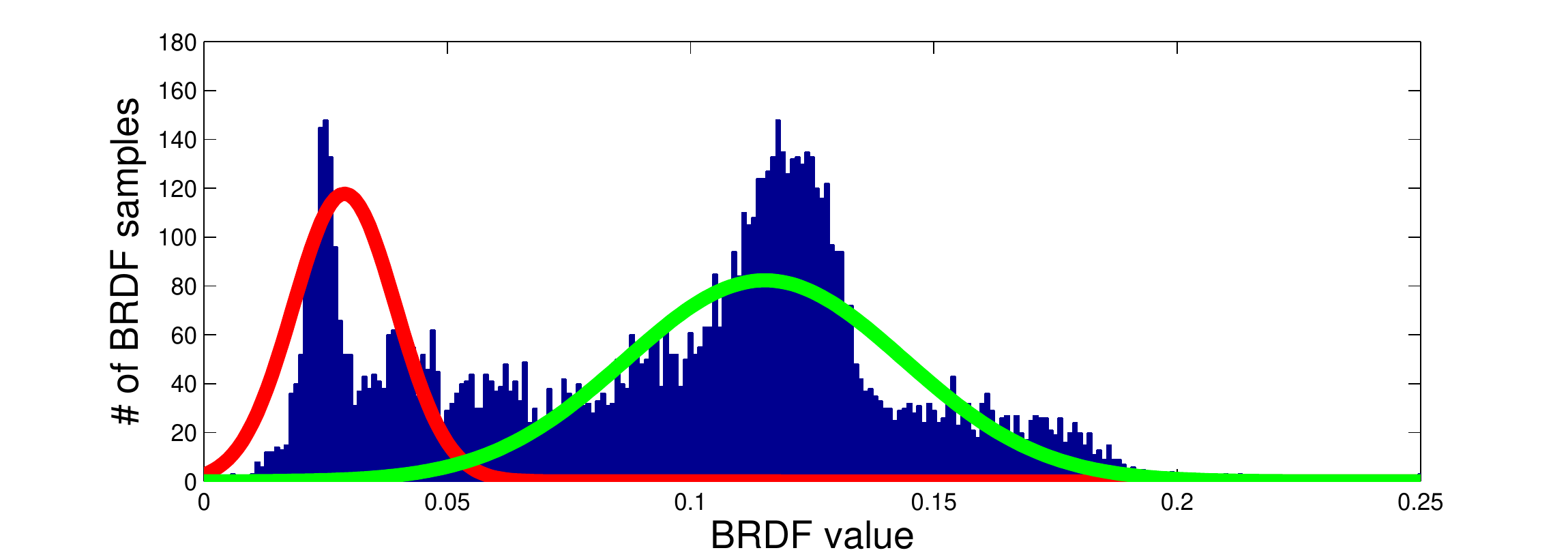}} \\ 
  (a) \\
  \raisebox{-0.5\height}{\includegraphics[width=\linewidth]{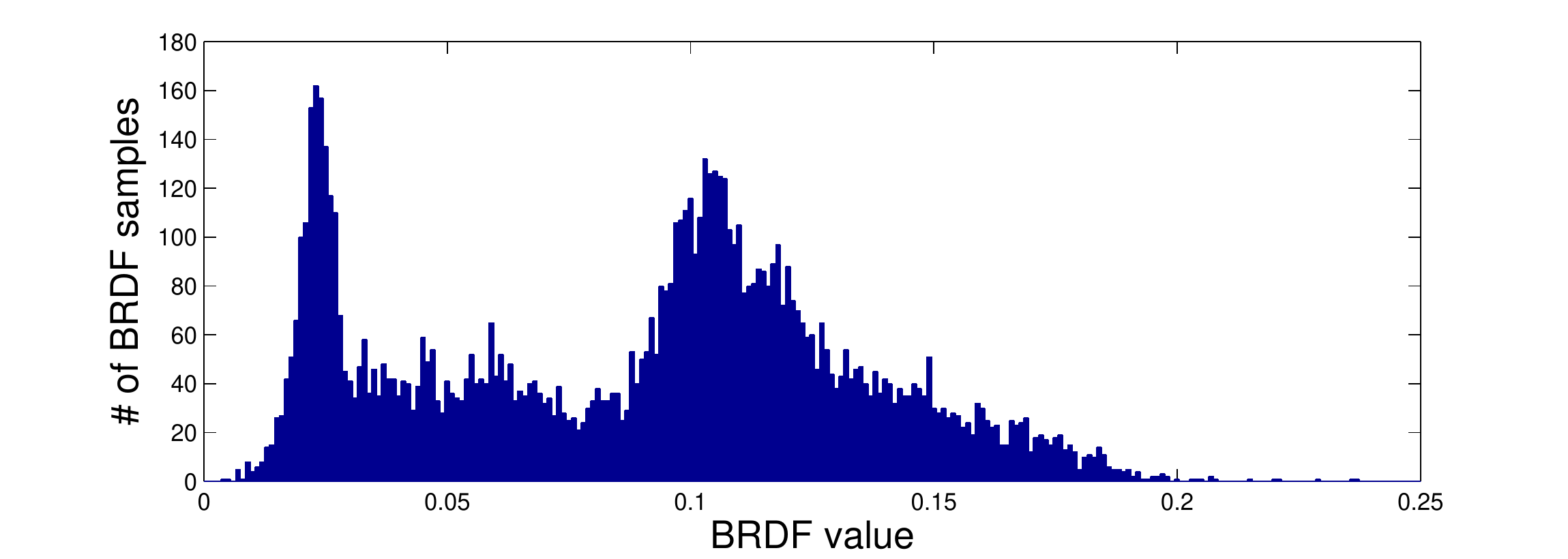}} \\
  (b) \\
  \raisebox{-0.5\height}{\includegraphics[width=\linewidth]{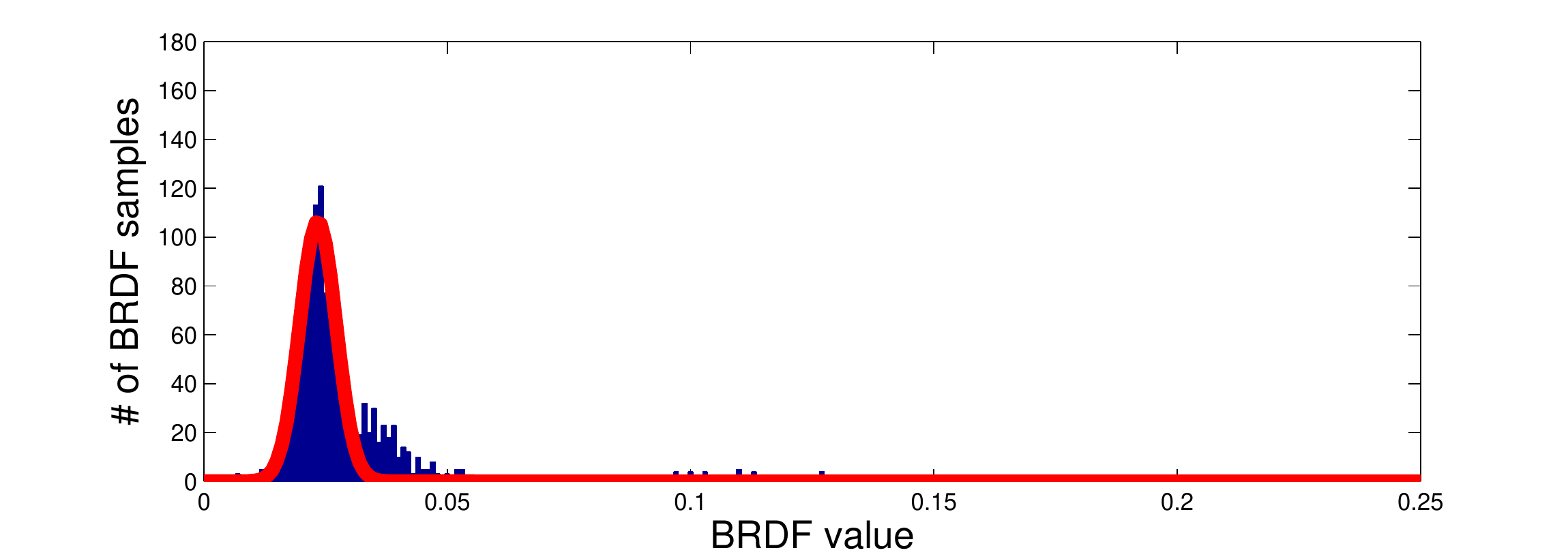}} \\
  (c) \\
  \raisebox{-0.5\height}{\includegraphics[width=\linewidth]{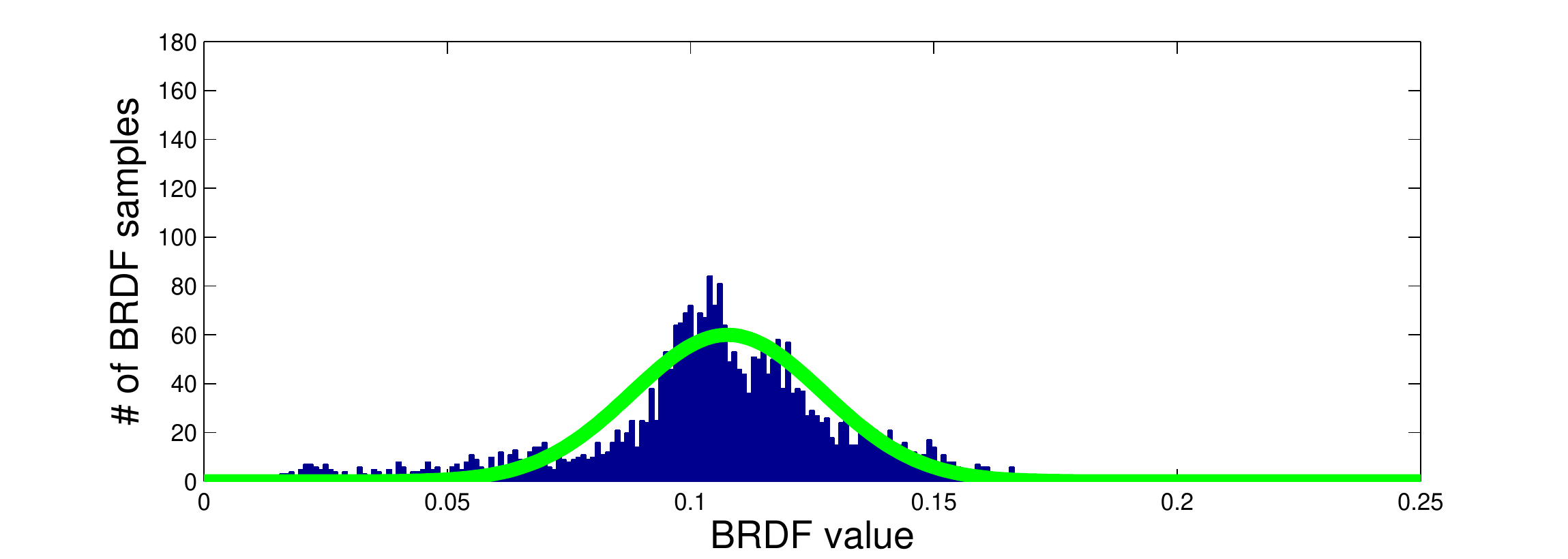}} \\
  (d)
\end{tabular}
\end{center}
\caption{1D illustration of algorithm. (a) shows the histogram of BRDF samples in the initially selected cell $C_1$. Two clusters in $C_1$ are fitted to Gaussians as shown by the red and green curves. These samples' corresponding vertices can be classified into two material groups $\mathrm{Mat}_1$ and $\mathrm{Mat}_2$ based on the two curves. (b) shows the histogram of another cell $C_2$ to be separated. (c)(d) show the histograms of BRDF samples in $C_2$ whose corresponding vertices belong to $\mathrm{Mat}_1$ and $\mathrm{Mat}_2$ respectively. Unclassified vertices associated with $C_2$ could be classified based on the curves in (c) and (d).}
\label{one_dim_illustration}
\vspace{-5mm}
\end{figure}

\noindent
\textbf{Line 2}
We choose the most `separable' cell in $T$ to initialize material segmentation.
We first model each cluster in every cell of $T$ by a Gaussian distribution(\figref{one_dim_illustration}(a)).
If there are two clusters in a BRDF cell, we define the following `separability score'
\begin{equation}
Score = \mathrm{MD}(\mu_2,\mu_1,S_1) + \mathrm{MD}(\mu_1,\mu_2,S_2)
\label{sep_score}
\end{equation}
where $\mu_{1}$, $\mu_{2}$, $S_{1}$ and $S_{2}$ are the mean vectors and covariance matrices for the two Gaussians respectively, and $\mathrm{MD}$ is Mahalanobis distance defined in the following way
\begin{equation}
\mathrm{MD}(x, \mu, S) = \sqrt{(x-\mu)S^{-1}(x-\mu)}.
\end{equation}
The cell with a single or no cluster has a score of zero.
The cell $\mathrm{Cell}_k$ in $T$ with the highest score is selected for initial segmentation.

\noindent
\textbf{Line 3}
For each sample in the selected cell $\mathrm{Cell}_k$, we calculate its Mahalanobis distances $\mathrm{MD}_1$ and $\mathrm{MD}_2$ to the two Gaussians. If $\mathrm{MD}_1 < 3$ and $\mathrm{MD}_2 > 3$, the sample's corresponding vertex will be placed into an empty vertex set $\mathrm{Mat}_1$. If $\mathrm{MD}_1 > 3$ and $\mathrm{MD}_2 < 3$, this vertex goes into another empty vertex set $\mathrm{Mat}_2$.
The two vertex sets correspond to two materials in the scene.
\figref{mat_segmentation}(b) shows $\mathrm{Mat}_1$ and $\mathrm{Mat}_2$ on the original mesh.

We note that value $3$ is chosen as threshold because it corresponds to the $3\sigma$ rule of a 1D Gaussian. In other words, if a sample is drawn from a Gaussian, there is a high probability that the Mahalanobis distance between them will be less than $3$.

\noindent
\textbf{Line 5}
In this step, we find a new cell in $T$ to be separated. The selection is based on a new separability score.

For cell $\mathrm{Cell}_i$, We denote the set of all its BRDF samples' corresponding vertices as $\mathrm{Set}_{i}$
 and find intersections $\mathrm{Set}_{i} \cap \mathrm{Mat}_1$ and $\mathrm{Set}_{i} \cap \mathrm{Mat}_2$. If $\mathrm{Set}_{i} \cap \mathrm{Mat}_1$(or $\mathrm{Set}_{i} \cap \mathrm{Mat}_2$) contains less than half of the vertices in $\mathrm{Mat}_1$(or $\mathrm{Mat}_2$), the new separability score of this BRDF cell will be set to zero. Otherwise, we estimate two Gaussian distributions in $\mathrm{Cell}_i$ with the vertices in $\mathrm{Set}_{i} \cap \mathrm{Mat}_1$ and $\mathrm{Set}_{i} \cap \mathrm{Mat}_2$. $\mathrm{Cell}_i$'s new separability score is calculated with the two new Gaussians using \equref{sep_score}. The cell in $T$ with the highest new separability score is selected.

\noindent
\textbf{Line 6}
We propagate the segmentation information of $\mathrm{Mat}_1$ and $\mathrm{Mat}_2$ to the selected cell $\mathrm{Cell}_l$.

From last step, we have already obtained two Gaussian distributions estimated from 
$\mathrm{Set}_{l} \cap \mathrm{Mat}_1$ and $\mathrm{Set}_{l} \cap \mathrm{Mat}_2$.
For every BRDF sample in $\mathrm{Cell}_l$ whose corresponding vertex does not belong to $\mathrm{Mat}_1 \cup \mathrm{Mat}_2$, we calculate its Mahalanobis distances $\mathrm{MD}_1$ and $\mathrm{MD}_2$ to the two Gaussians of $\mathrm{Set}_{i} \cap \mathrm{Mat}_1$ and $\mathrm{Set}_{i} \cap \mathrm{Mat}_2$. If $\mathrm{MD}_1 < 3$ and $\mathrm{MD}_2 > 3$, the vertex will be added to $\mathrm{Mat}_1$. If $\mathrm{MD}_1 > 3$ and $\mathrm{MD}_2 < 3$, it goes into $\mathrm{Mat}_2$. See \figref{one_dim_illustration}.

During this step, the material segmentation information encoded in $\mathrm{Mat}_1$ and $\mathrm{Mat}_2$ is propagated to $\mathrm{Cell}_l$, which in turn improves the segmentation by adding additional vertices into $\mathrm{Mat}_1$ and $\mathrm{Mat}_2$.

\mycomment{
\textbf{Selection of Next BRDF Cell} and \textbf{Segmentation Propagation} are repeated until there is no BRDF cell which can be segmented, either because of being already segmented or because there are less than two clusters in the BRDF cell.
}

\figref{mat_segmentation}(c-f) shows how sets $\mathrm{Mat}_1$ and $\mathrm{Mat}_2$ grows as the above  algorithm proceeds.

Due to data noise and incomplete BRDF tables, some vertices are never classified into the either $\mathrm{Mat}_1$ or $\mathrm{Mat}_2$. In this case, we do not force such vertices into either set because there is a lack of information for us to make confident assertions about them.

\noindent
\textbf{Multi-Material Extension}
In the case of multi-material segmentation, the meanshift algorithm in \textbf{Step 1} could produce over 2 clusters in a cell of $T$. Then separability can be defined for each \textit{cluster} as its Mahalanobis distance to the `nearest' cluster within the same cell. The segmentation will start from the most separable cluster and be propagated in every cell in $T$. After each round, a new material is segmented out. This kind of segmentation could proceed till no new material is identified. Clearly, the user does not need to specify the number of materials in advance.

\vspace{-5mm}
\subsubsection{Post-processing}
After identifying vertices of the same material, we can merge their associated BRDF tables and obtain a more complete and accurate representation of the material's BRDF. 
Even so, however, there are still empty cells in the BRDF table. This is mainly due to two reasons. First, we discard BRDF values observed when $\omega_{in} > 60^o$ or $\omega_{out} > 60^o$. Second, because of the fixed baseline between the IR camera and LEDs, $\theta_d$ only varies between $0^o$ and $20^o$ and there is no BRDF samples beyond this range.
Thus, our system can only provide a portion of the full BRDF.
While we could fit the available BRDF samples to either a parametric model or a set of measured real-world BRDFs as~\cite{noll2014robust}, we adopt the simple approach of linear interpolation and extrapolation to complete the BRDF table. Note that this completion is only for rendering purpose and should not be considered as valid measurement.

\section{Experiments}

\begin{figure*}
\begin{center}
\addtolength{\tabcolsep}{-4pt}  
\begin{tabular}{cccc}
  \raisebox{-0.5\height}{\includegraphics[width=0.25\linewidth]{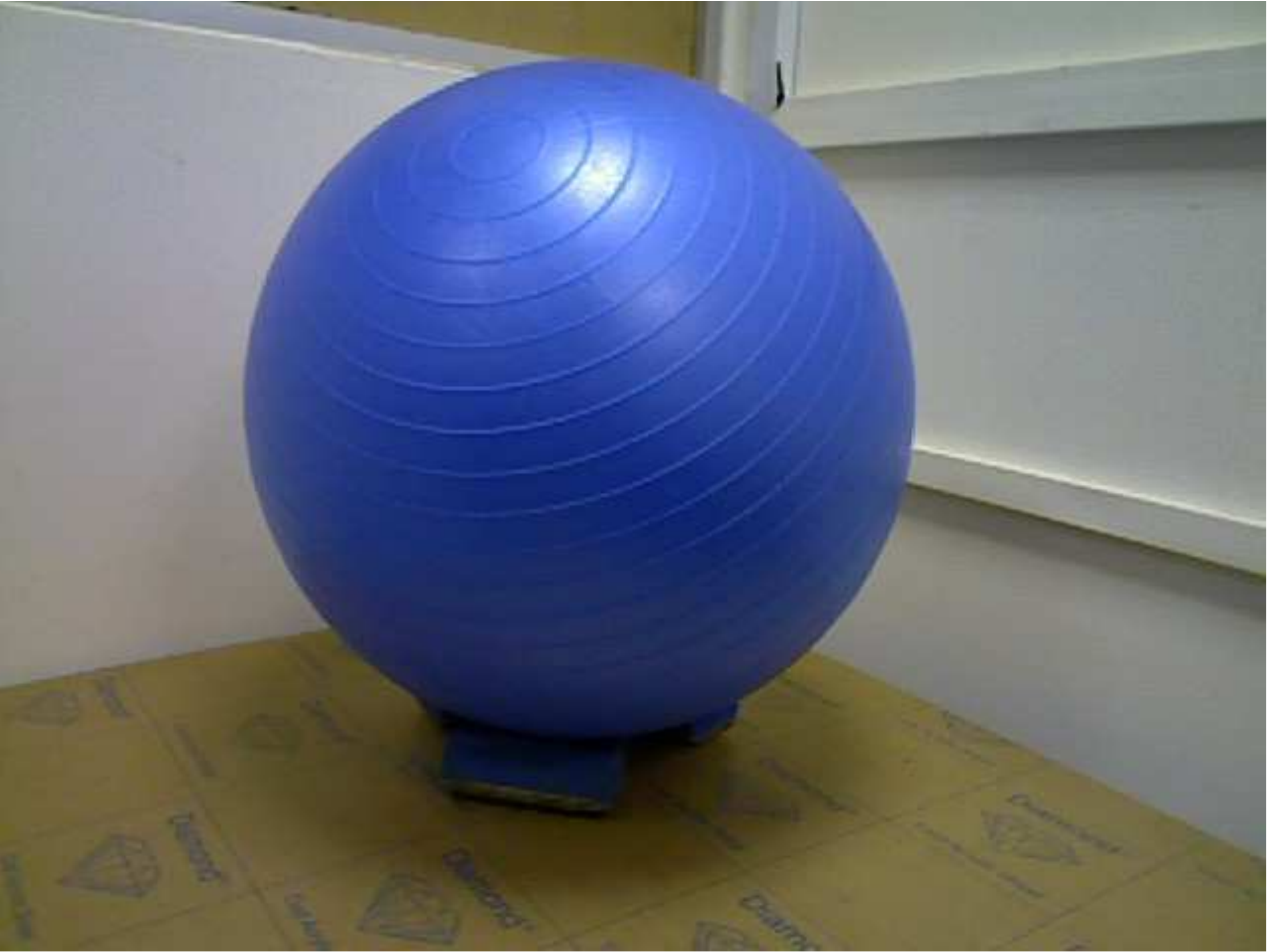}} & \raisebox{-0.5\height}{\includegraphics[width=0.25\linewidth]{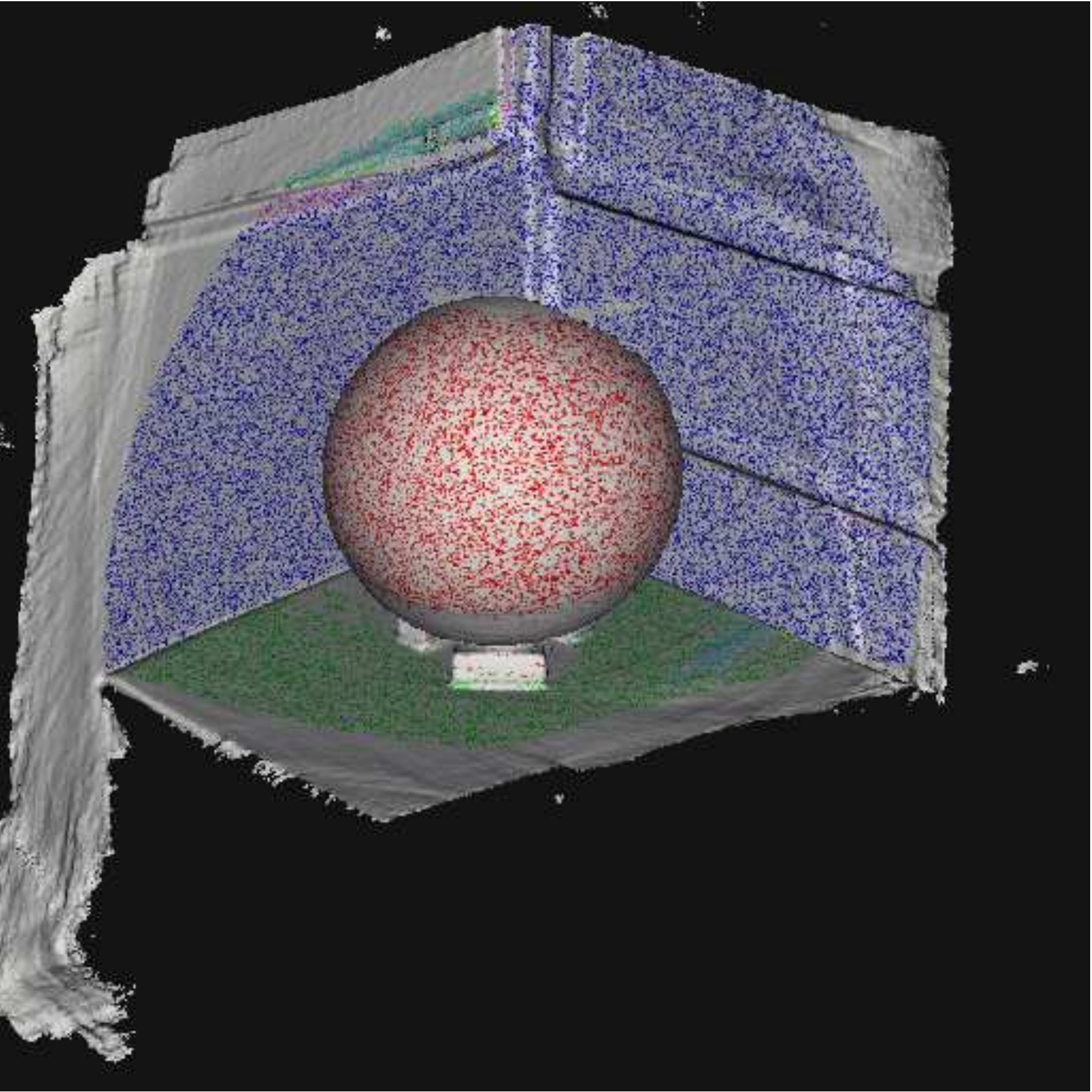}}
  & \raisebox{-0.5\height}{\includegraphics[width=0.25\linewidth]{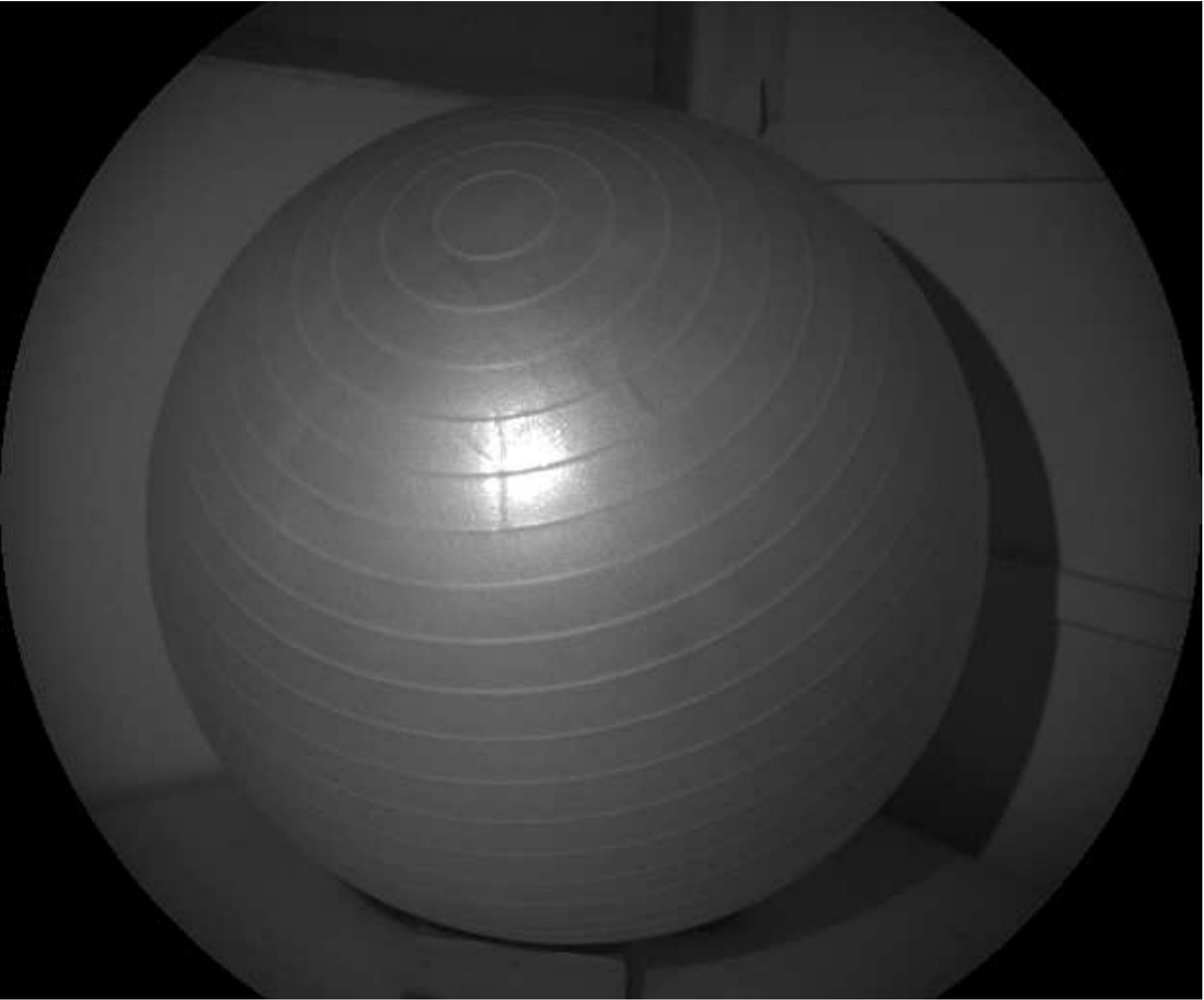}} & \raisebox{-0.5\height}{\includegraphics[width=0.25\linewidth]{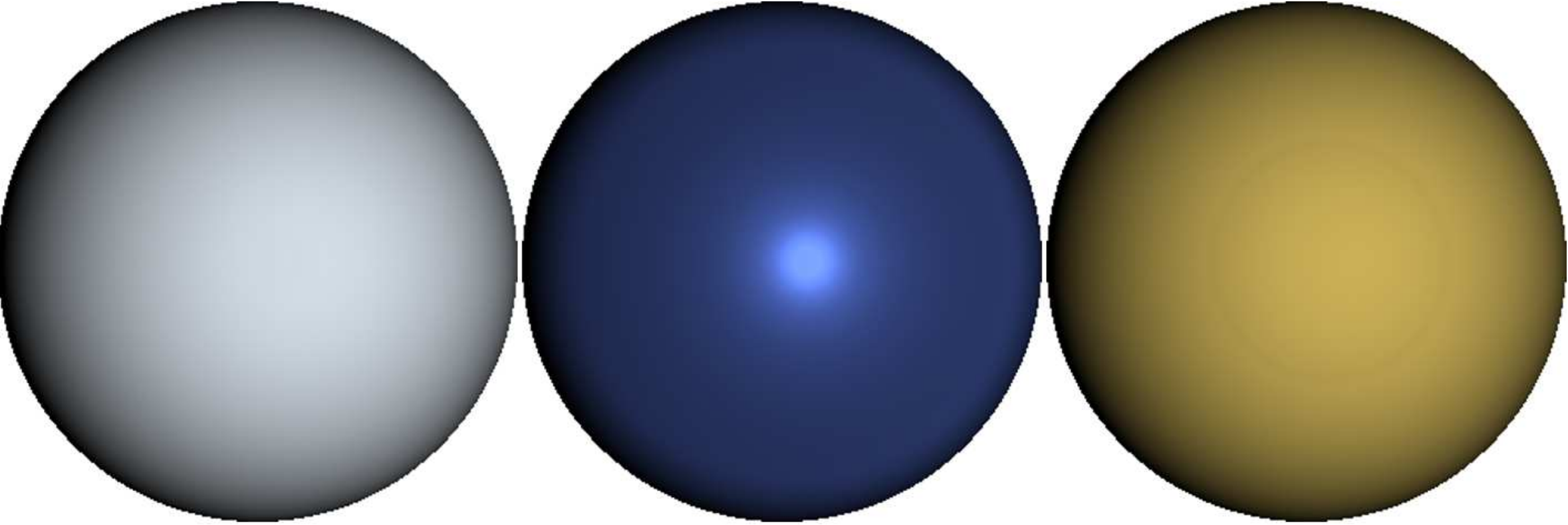}} \\
(a) & (c) & (e) & (g) \\
	\raisebox{-0.5\height}{\includegraphics[width=0.25\linewidth]{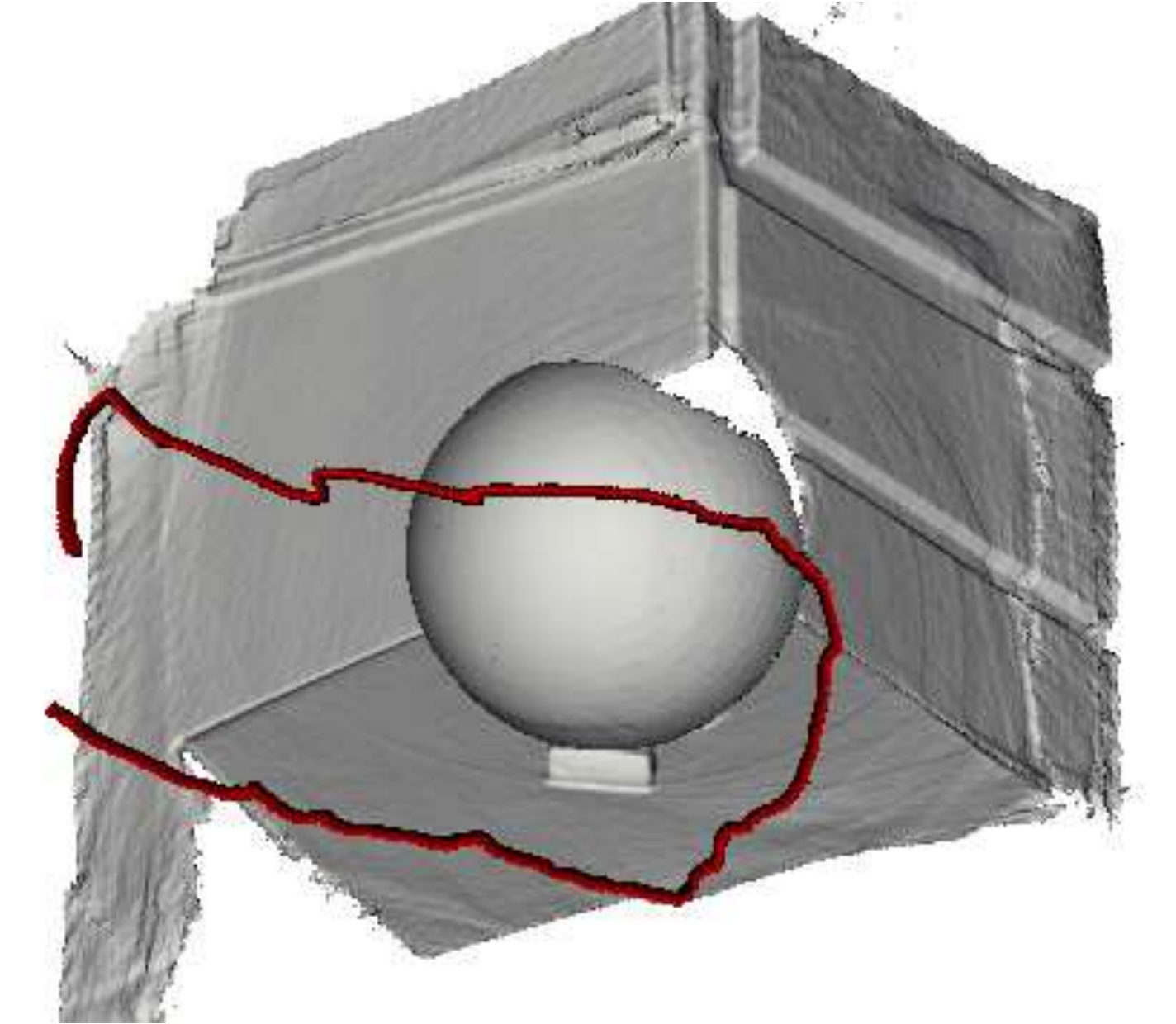}} & \raisebox{-0.5\height}{\includegraphics[width=0.25\linewidth]{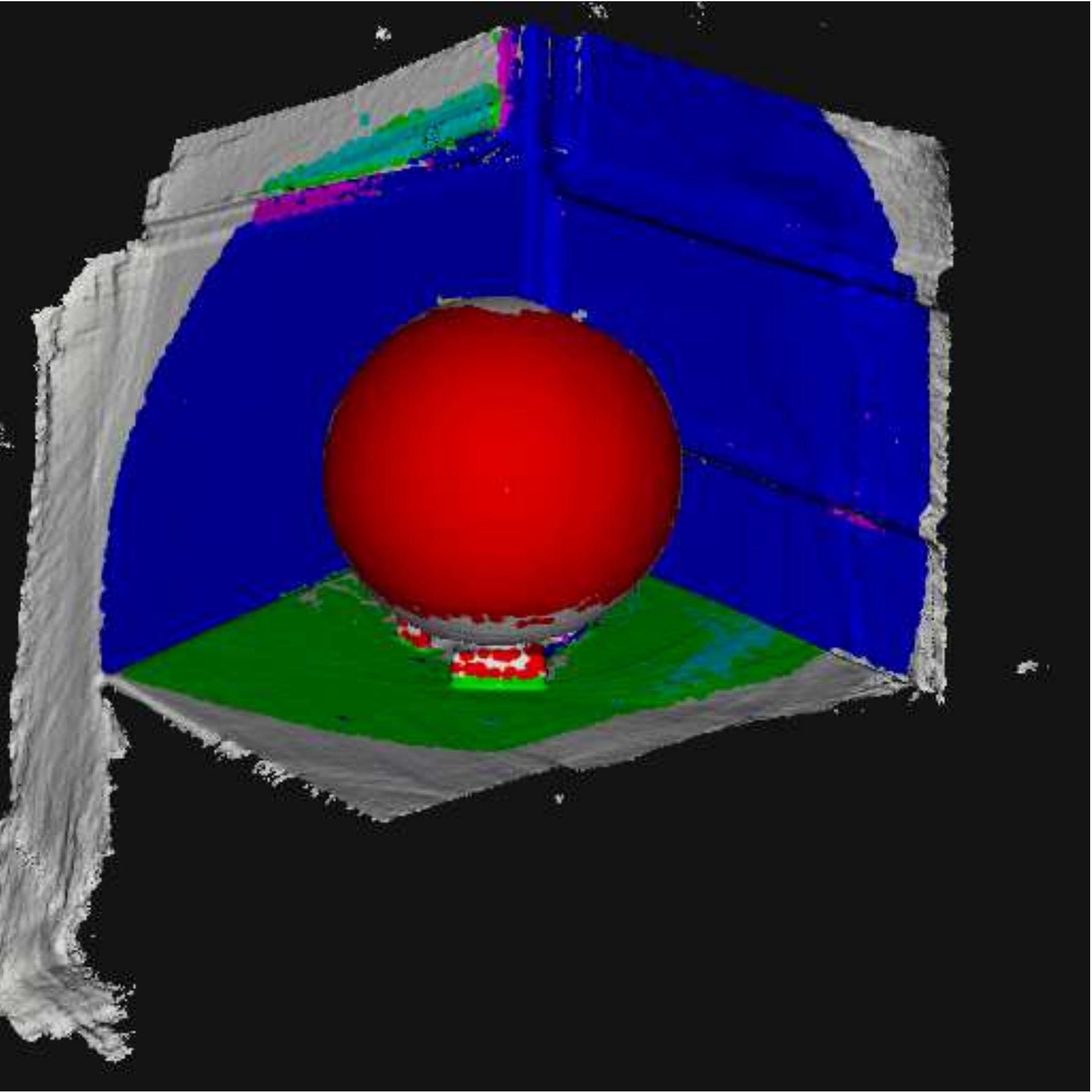}} & \raisebox{-0.5\height}{\includegraphics[width=0.25\linewidth]{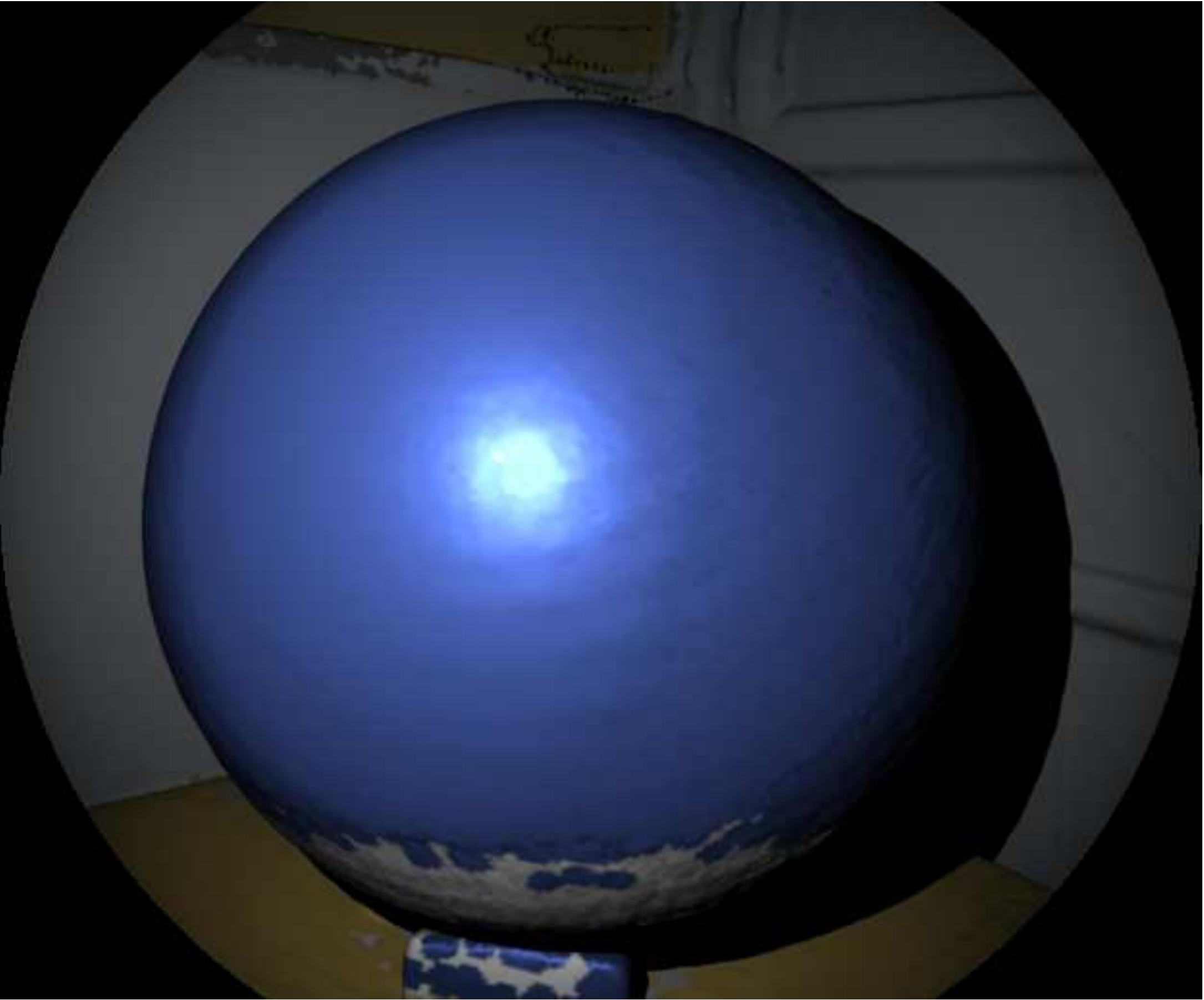}} & \raisebox{-0.5\height}{\includegraphics[width=0.25\linewidth]{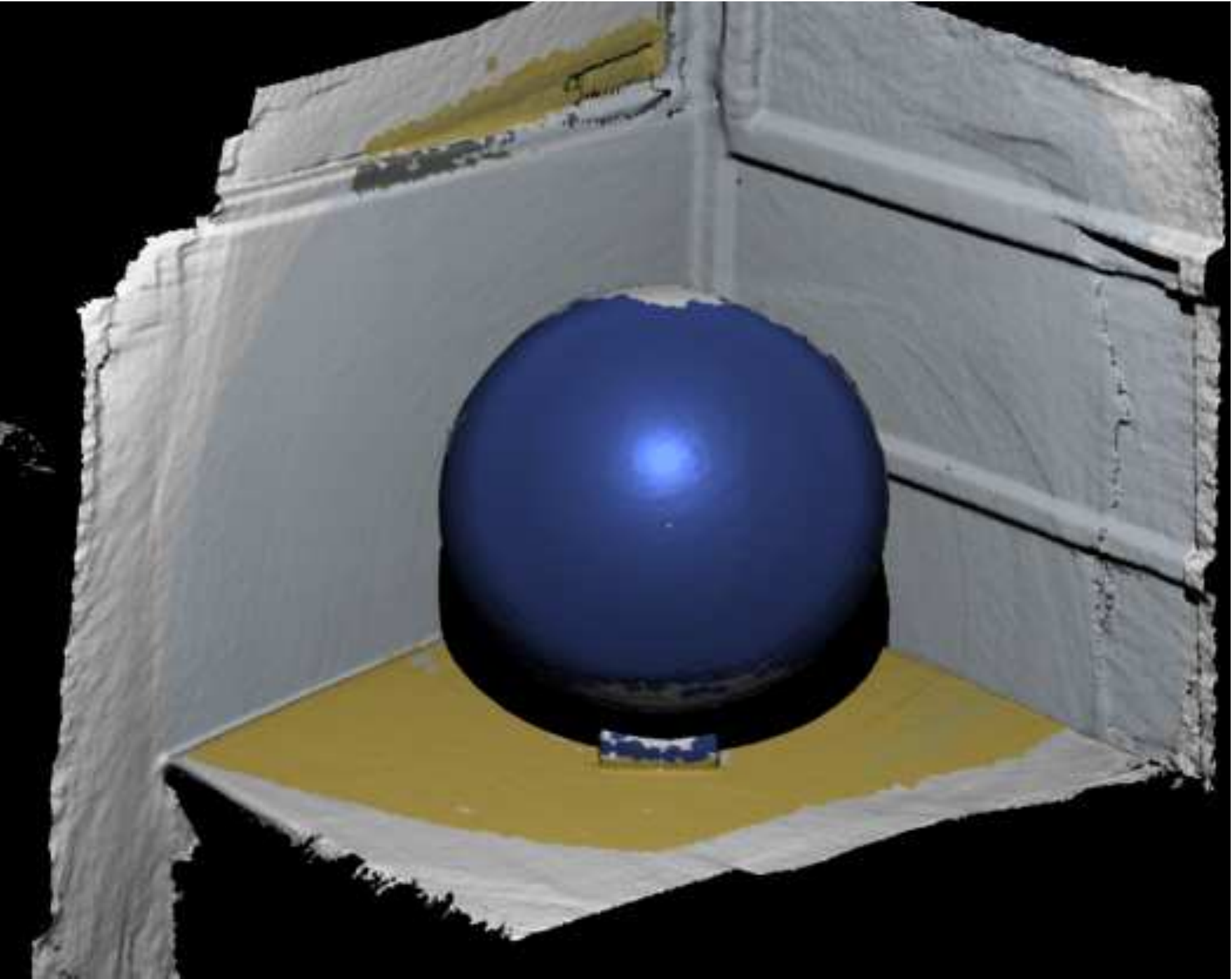}} \\
(b) & (d) & (f) & (h)
\end{tabular}
\addtolength{\tabcolsep}{4pt}  
\end{center}
\caption{(a) shows an input RGB image. (b) shows the mesh and camera trajectory.(c-d) are material segmentation results: (c) is the results of 100,000 vertices while (d) is the final results. Note that not all vertices are classified into a material group. (e-f) are a pair of original IR image and a rendering under the same condition using estimated BRDFs. 
(g) shows the rendering of 3 dominant materials: wall, gymball and the board under gymball. (h) shows a novel rendering of the scene under novel lighting.}
\label{gymball_results}
\vspace{-5mm}
\end{figure*}

We demonstrate our system by results on three real-world scenes. Note that triangular meshes were produced by KinectFusion
without any manual cropping.

\figref{gymball_results} shows results on the scene \textit{Gymball}. (a) is an input RGB image
and (b) shows the reconstructed 3D shape as well as the depth camera's trajectory produced by KinectFusion.

Our material segmentation algorithm in \secref{set:material_segmentation} could be slow because of meanshift clustering during initialization.
Thus, we ran our algorithm on a set of 100,000 randomly sampled vertices instead of all vertices in the scene.
(c) shows the color-coded segmentation result of the sampled vertices. For those unsampled vertices, we simply
assume that each of them is of the same material as the nearest sampled vertex within $1cm$'s distance. (d) shows
this final segmentation result of all vertices. Please note that many vertices(in white) near the boundary of the scene
are not classified to any material group due to lack of valid BRDF observations or large discrepancy from other vertices' BRDF observations.

In total, our system generated 6 material groups and the number of vertices in each group is 7450, 33089, 8655, 1389, 269 and 140. The three largest material groups corresponded to the three dominant materials in the scene: the rubber-made gymball, the wall and the board beneath the gymball. The rest 3 material groups either corresponded to less dominant materials, or were due to various noise and large errors of surface normals. we rendered the three remaining materials in(g), where a sphere is rendered with different BRDFs under directional illumination. It is obvious that our system successfully captured the highlight of the gymball.
(e) shows one of the input IR image and (f) is the rendering of the scene under the same condition as (e).
Clearly, (f) is a colored version of (e). This resemblance suggests success of our reflectance estimation. (h) is a rendering of the scene under a novel lighting/viewing condition.

Please note that as mentioned earlier, some vertices are not associated with any materials. For these vertices, we assume Lambertian BRDF model to facilitate rendering. Also note that we mask out the four corners of (e) and (f) because we did not use these regions for strong vignette effect.

\begin{figure*}
\begin{center}
\addtolength{\tabcolsep}{-4pt}  
\begin{tabular}{cccc}
  \raisebox{-0.5\height}{\includegraphics[width=0.25\linewidth]{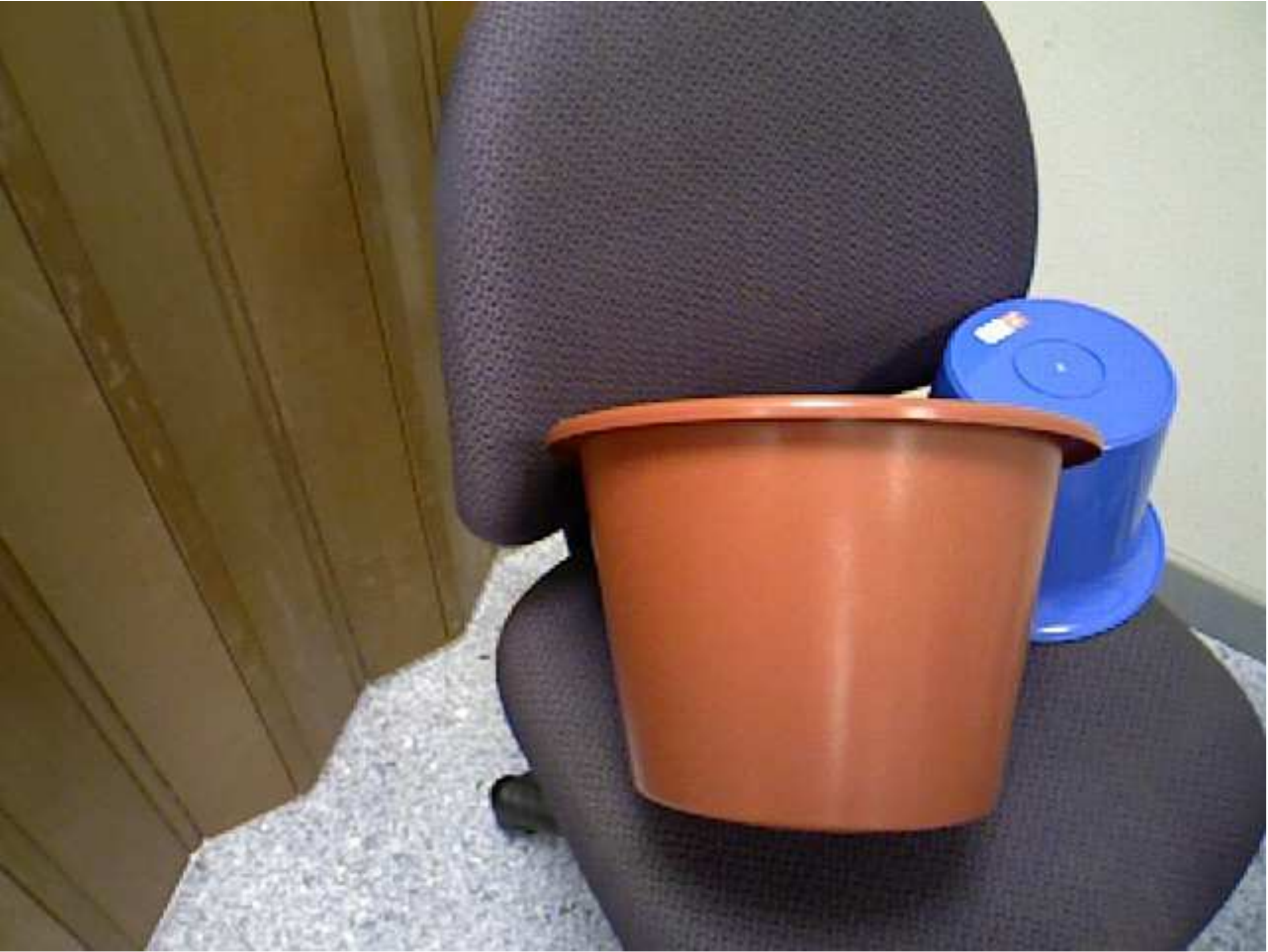}} & \raisebox{-0.5\height}{\includegraphics[width=0.25\linewidth]{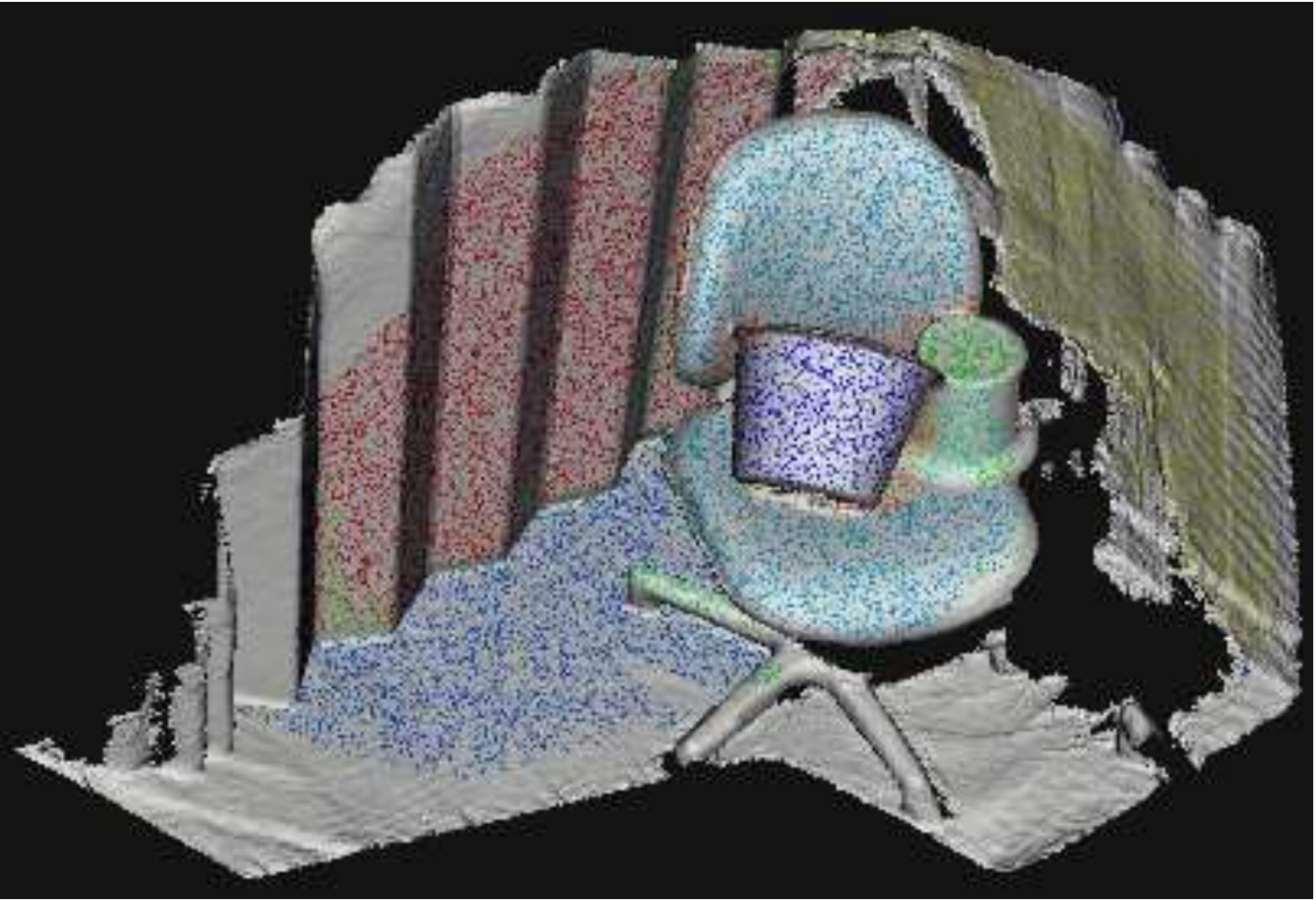}}
  & \raisebox{-0.5\height}{\includegraphics[width=0.25\linewidth]{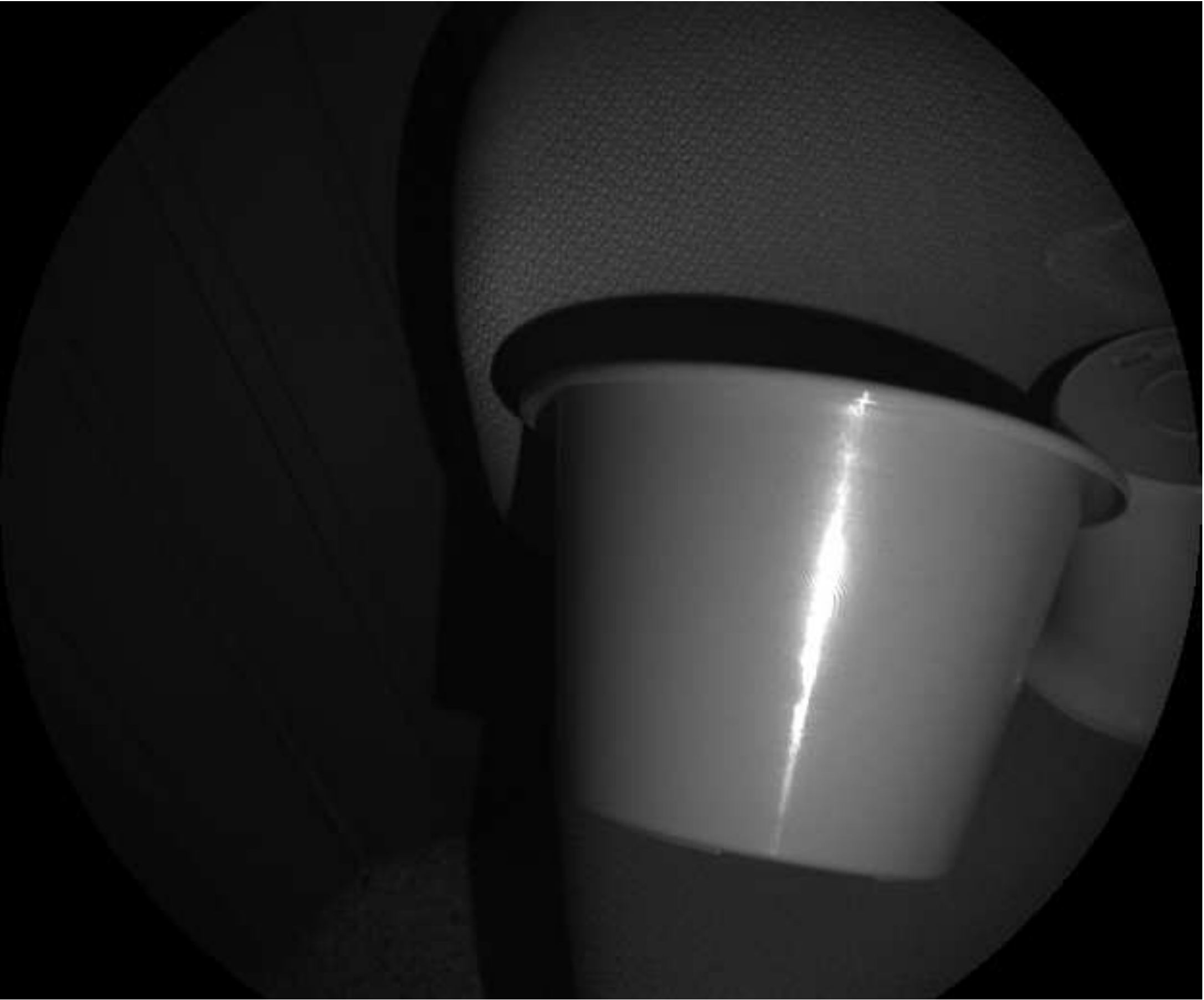}} & \raisebox{-0.5\height}{\includegraphics[width=0.25\linewidth]{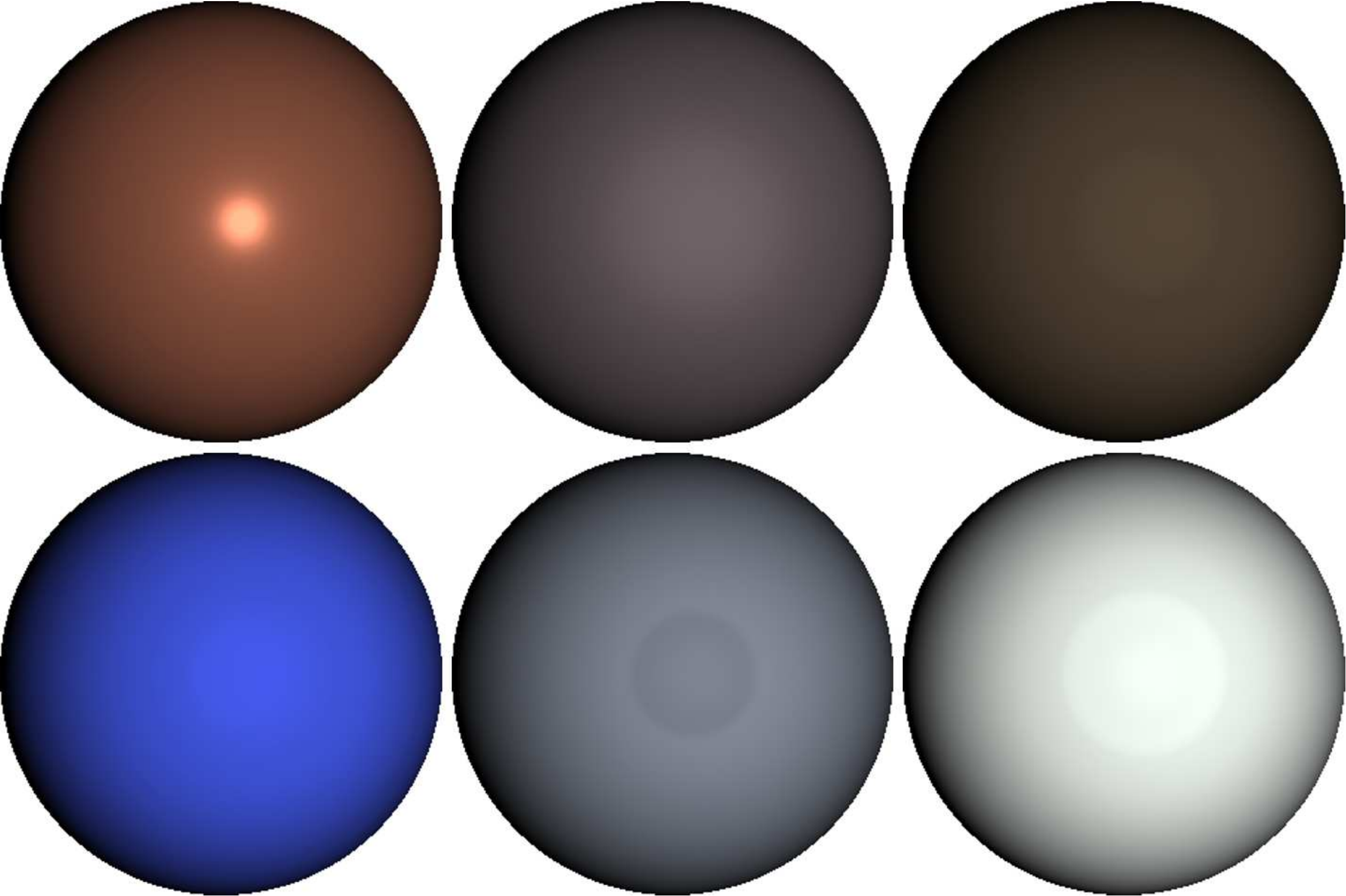}} \\
(a) & (c) & (e) & (g) \\
	\raisebox{-0.5\height}{\includegraphics[width=0.25\linewidth]{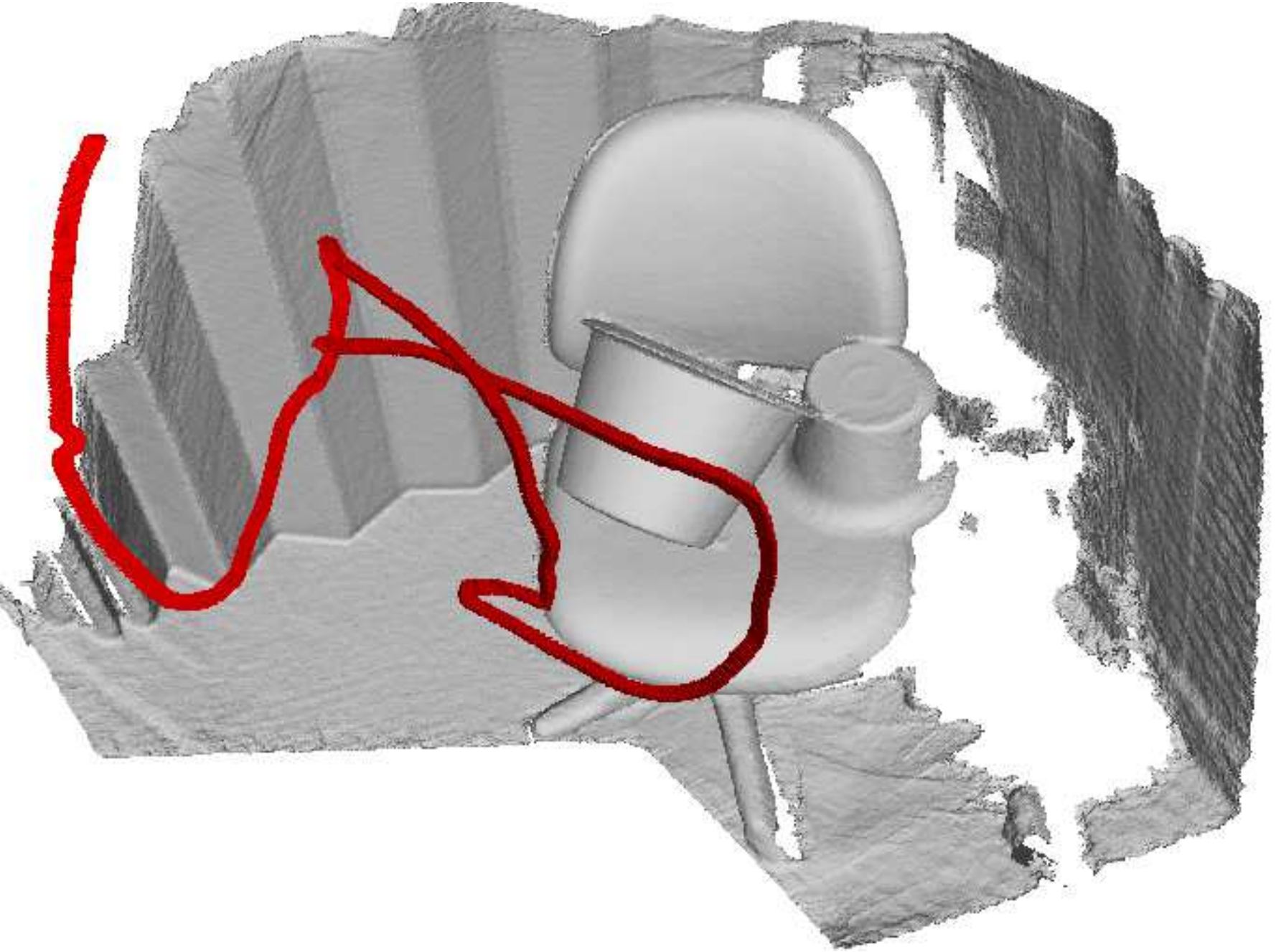}} & \raisebox{-0.5\height}{\includegraphics[width=0.25\linewidth]{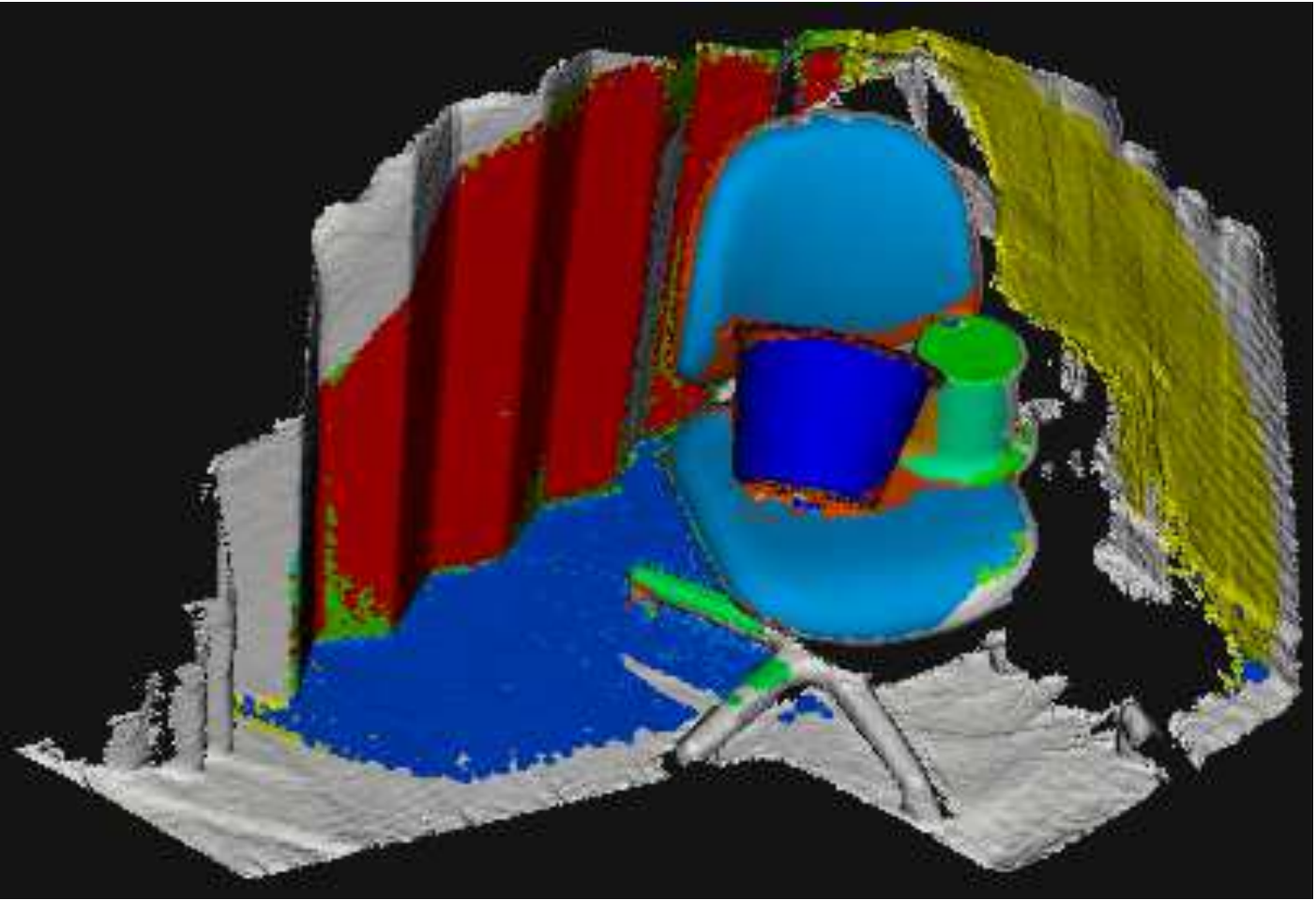}} & \raisebox{-0.5\height}{\includegraphics[width=0.25\linewidth]{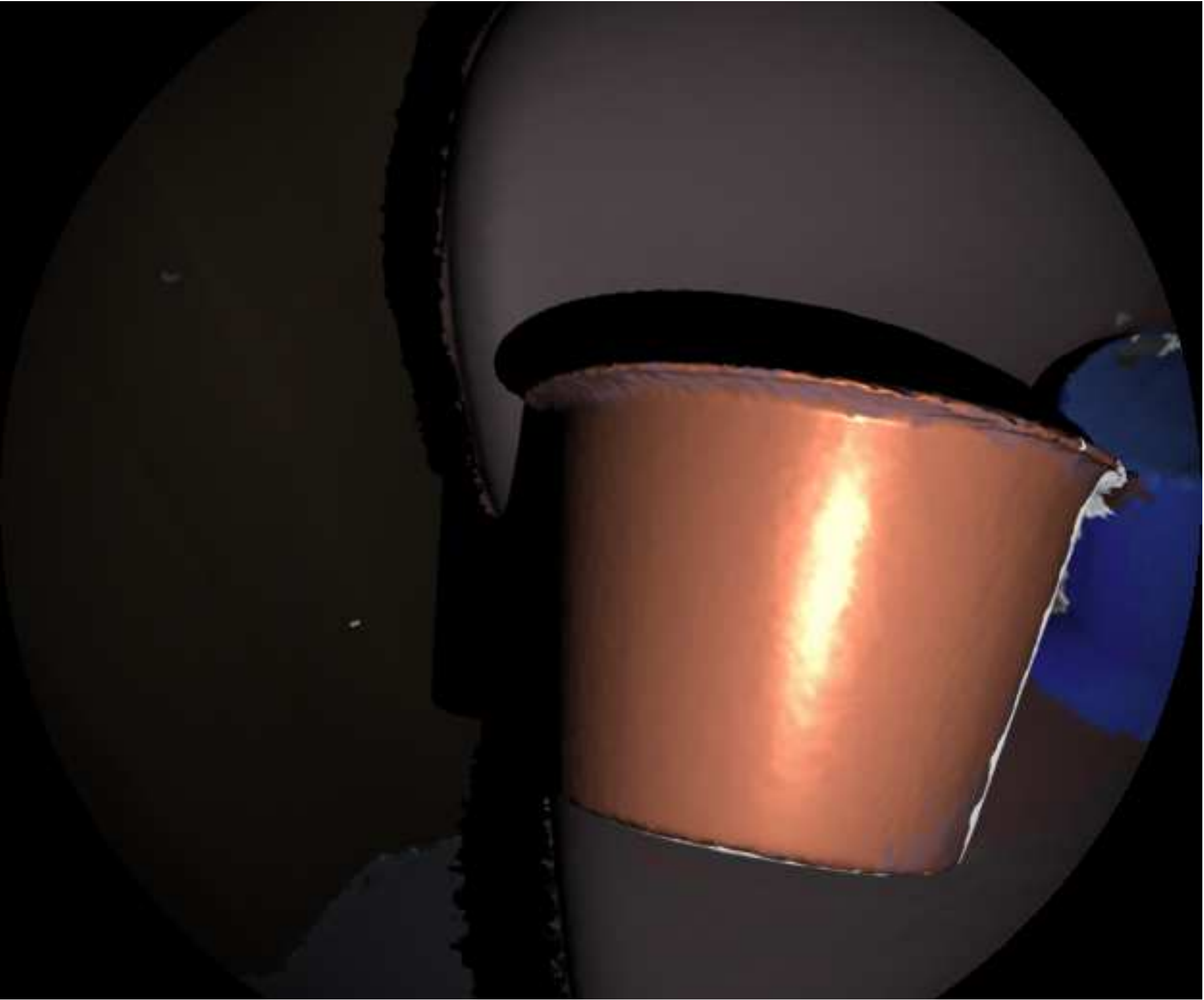}} & \raisebox{-0.5\height}{\includegraphics[width=0.25\linewidth]{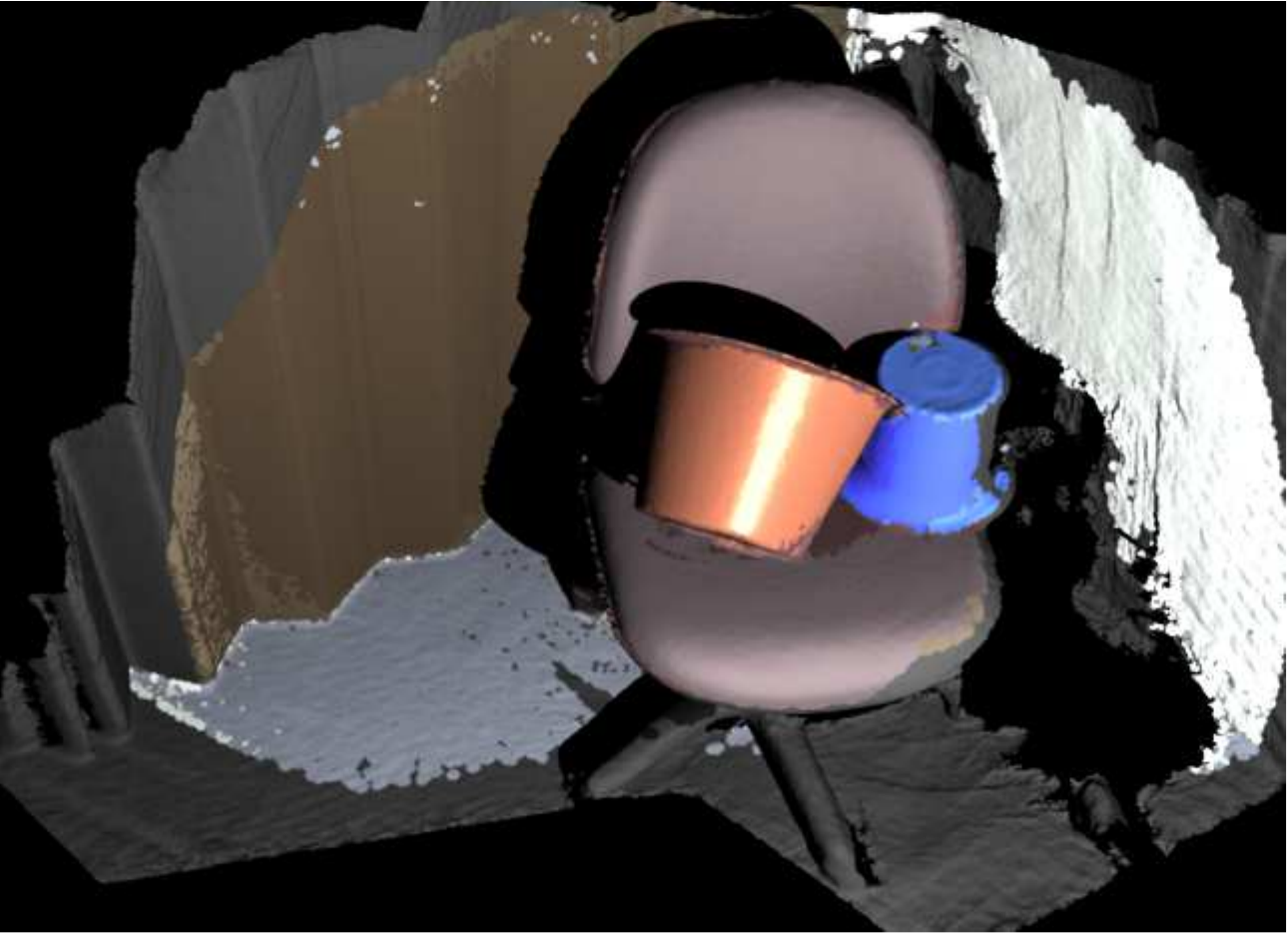}} \\
(b) & (d) & (f) & (h)
\end{tabular}`
\addtolength{\tabcolsep}{4pt}  
\end{center}
\caption{This figure shows results of the scene \textit{Pot}. Arrangement of subfigures is similar to \figref{gymball_results}. Note that the 6 materials in (g) correspond to the brown pot, chair, curtain, blue pot, ground and wall respectively. }
\label{pot_results}
\vspace{-2mm}
\end{figure*}

\figref{pot_results} shows results of another scene \textit{Pot}. As can be seen, different materials in the scene are well
segmented by our algorithm and the renderings of a sphere and the original scene demonstrate successful capture of highlight
observed in the brown plastic pot. However, our system failed to capture the highlight of the blue pot, which is also made of plastic. This is because our sampling of vertices was too sparse to observe highlight on the blue pot. 
For this scene, our system produces 19 material groups, 6 out of which are dominant. (g) shows the rendering of the 6
dominant materials, which correspond respectively to brown pot, chair, curtain, blue pot, ground and wall.

\begin{figure*}
\begin{center}
\addtolength{\tabcolsep}{-4pt}  
\begin{tabular}{cccc}
  \raisebox{-0.5\height}{\includegraphics[width=0.25\linewidth]{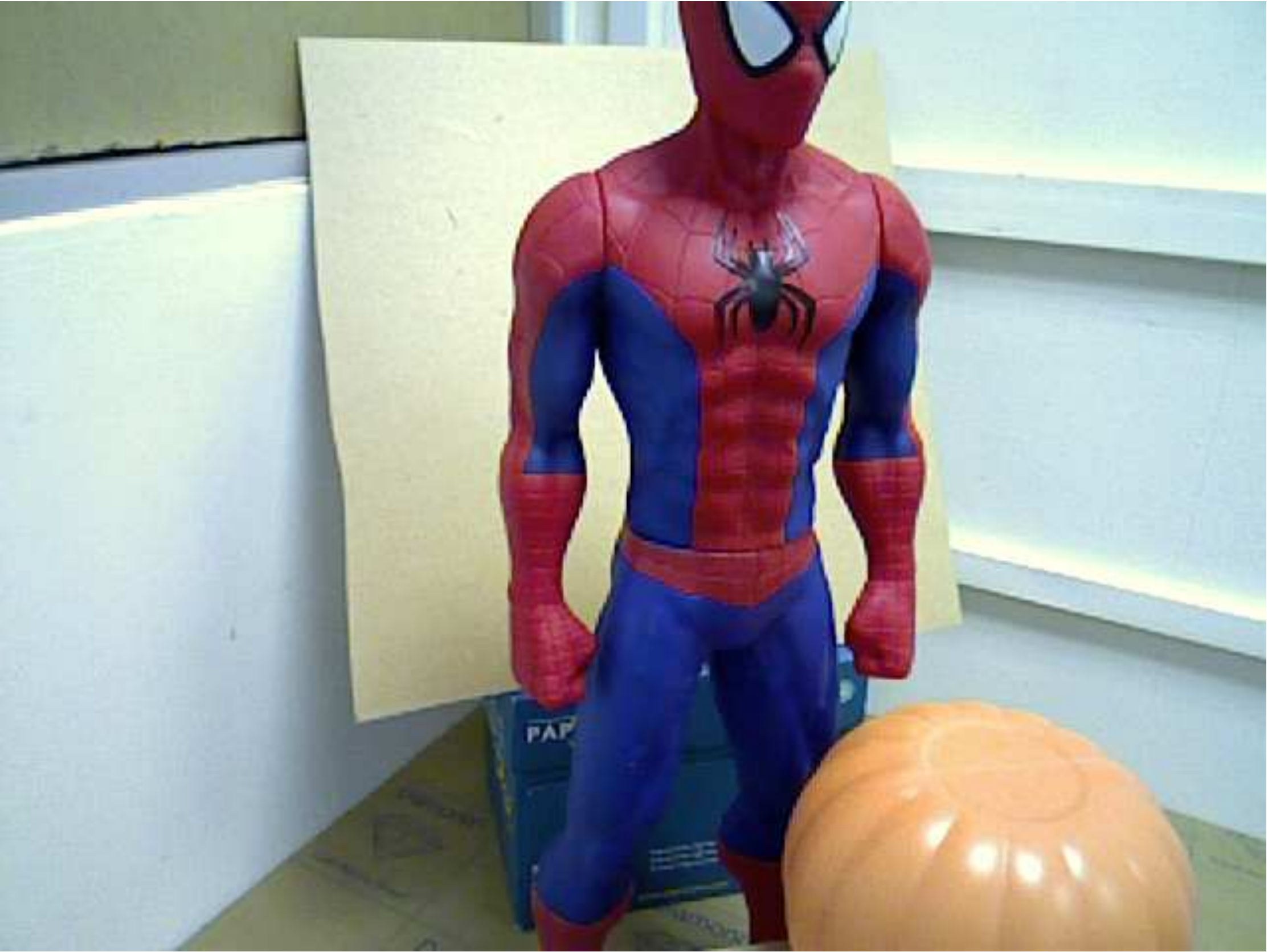}} & \raisebox{-0.5\height}{\includegraphics[width=0.25\linewidth]{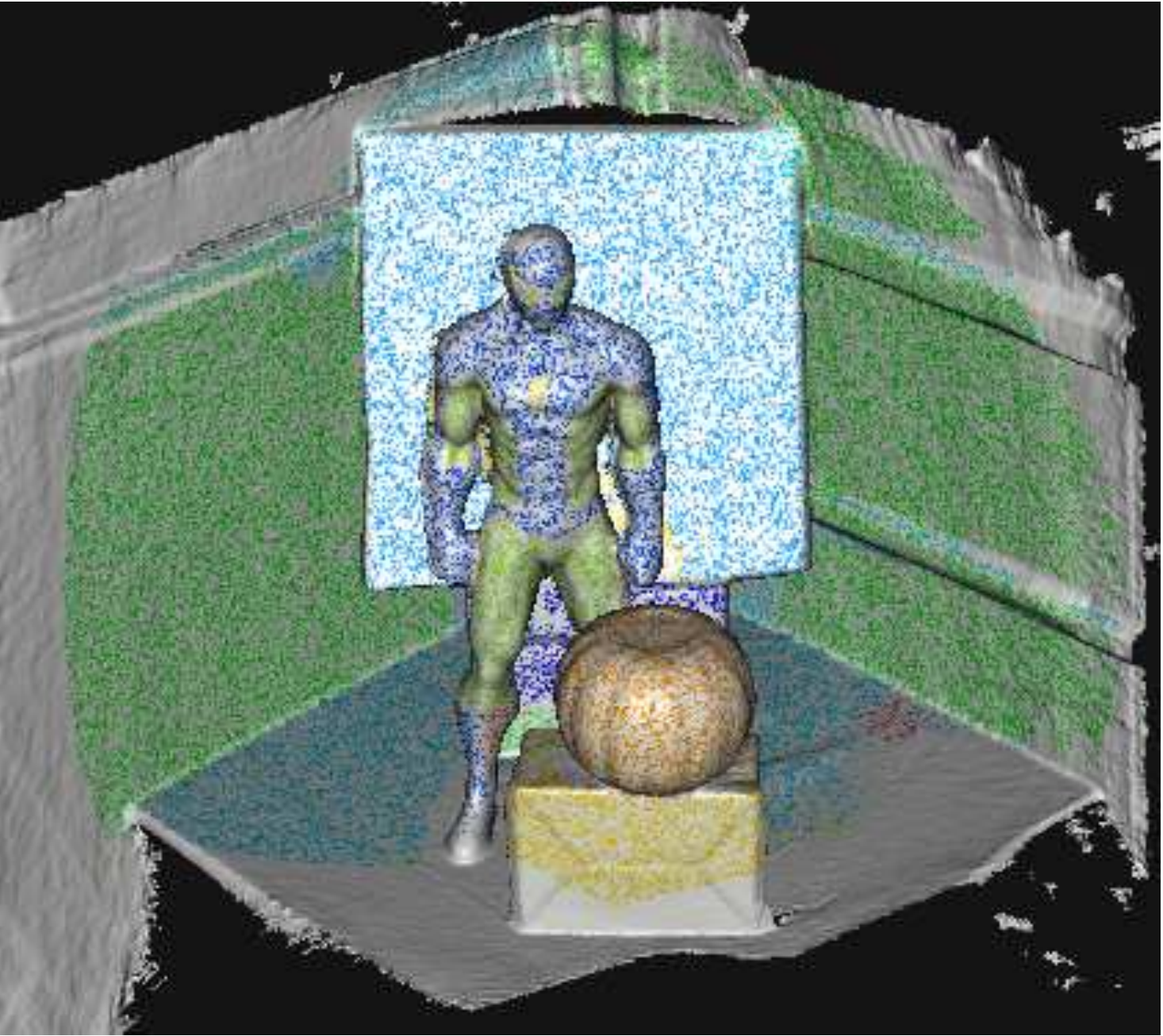}}
  & \raisebox{-0.5\height}{\includegraphics[width=0.25\linewidth]{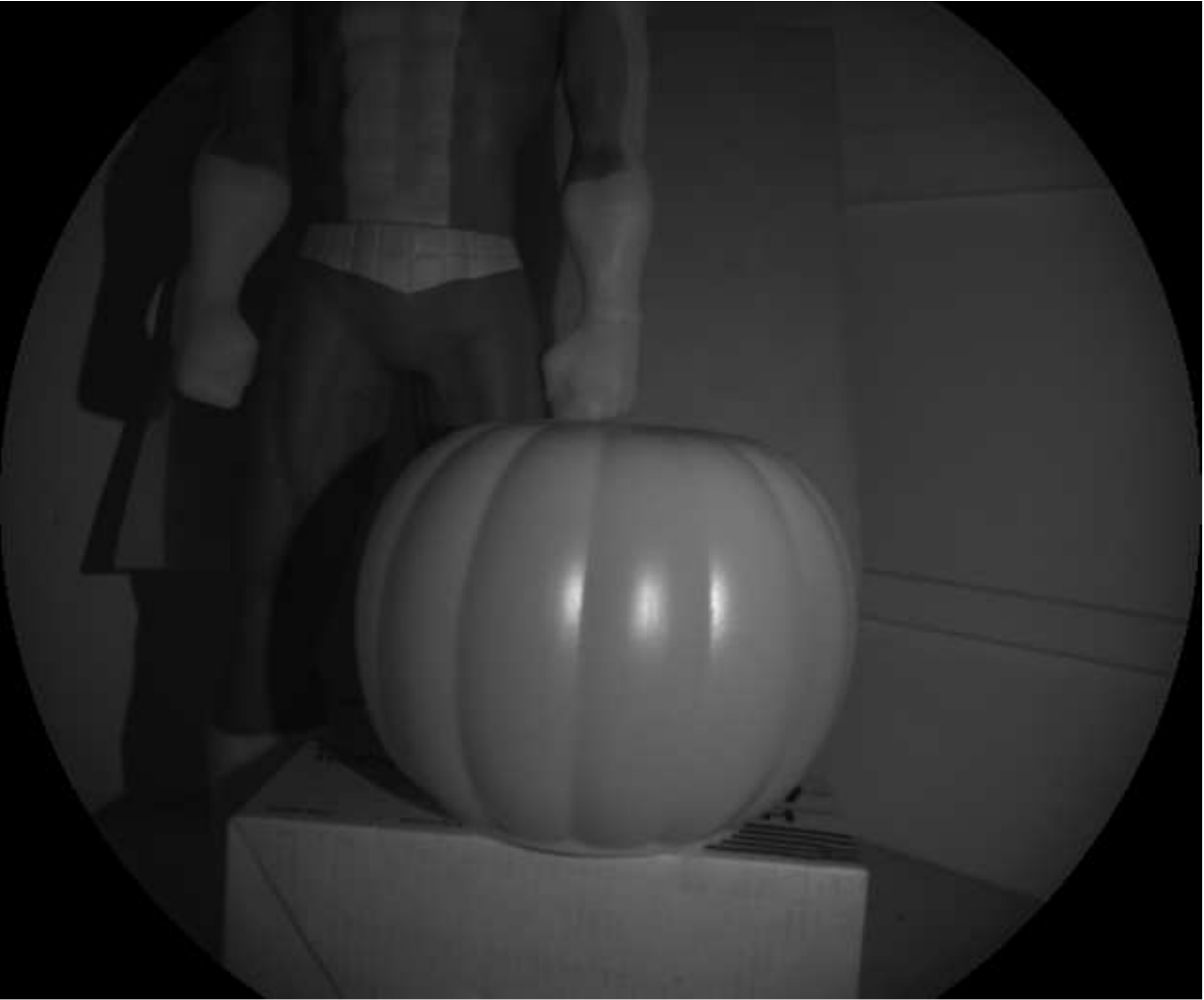}} & \raisebox{-0.5\height}{\includegraphics[width=0.25\linewidth]{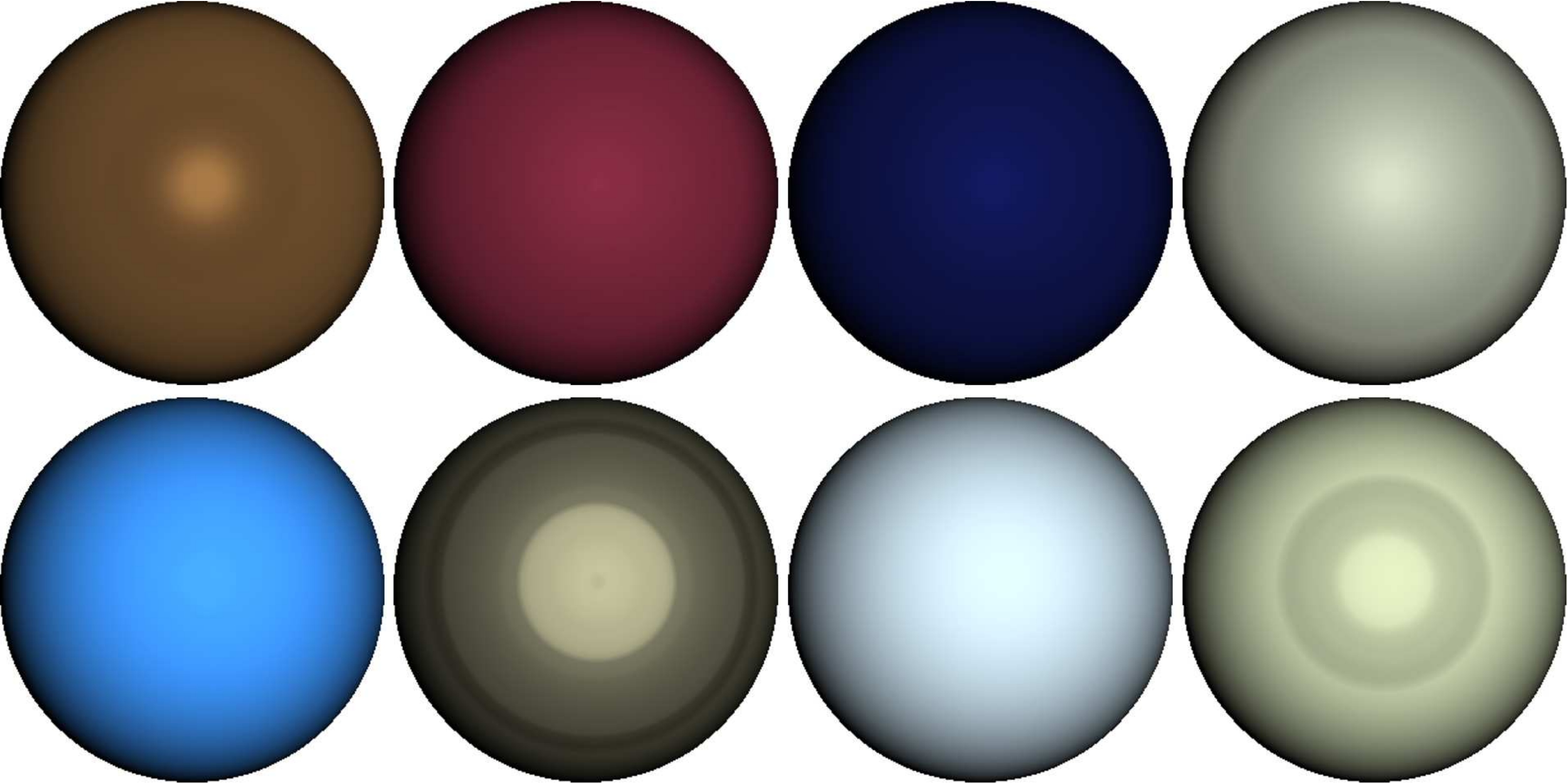}} \\
(a) & (c) & (e) & (g) \\
	\raisebox{-0.5\height}{\includegraphics[width=0.25\linewidth]{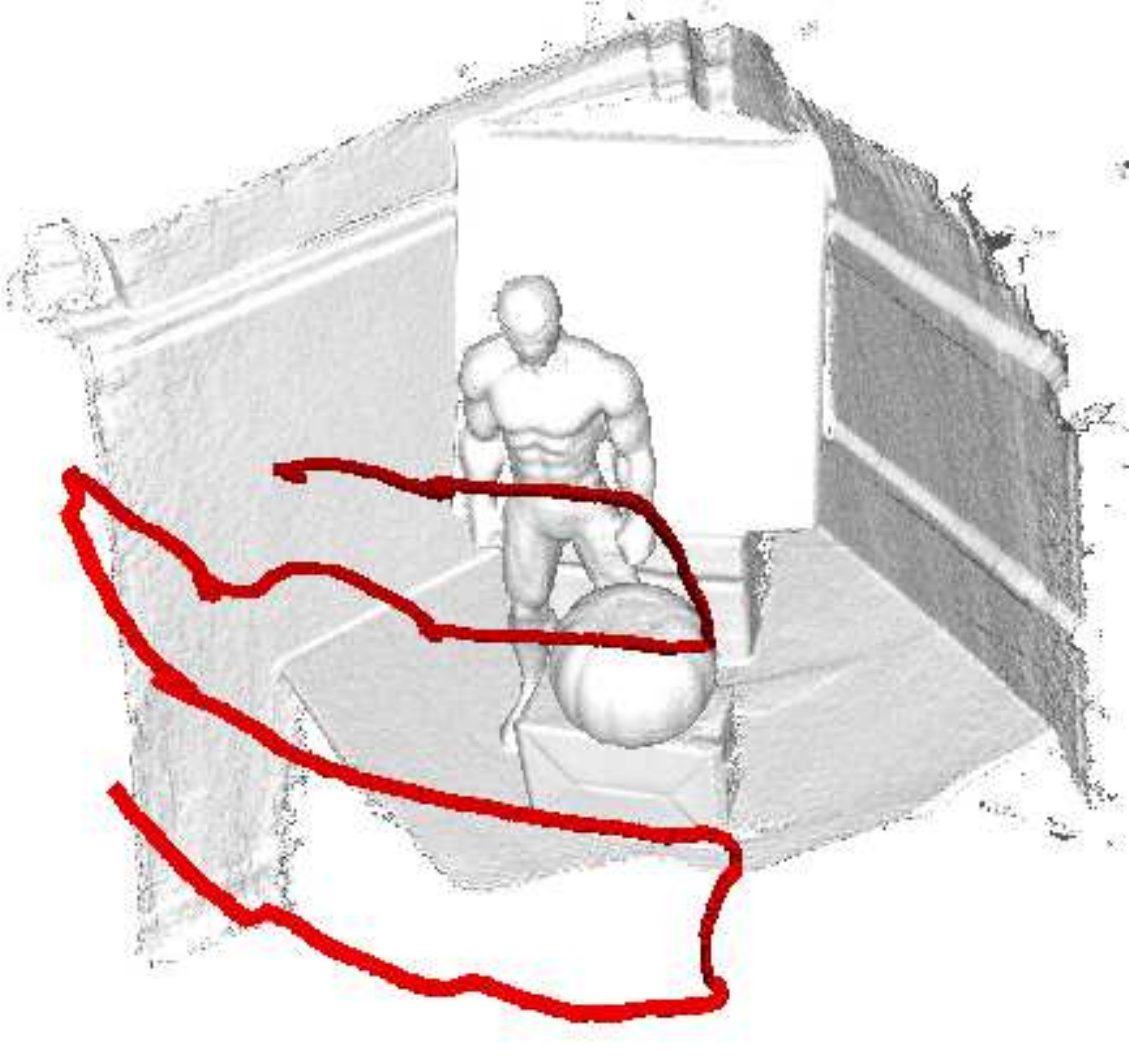}} & \raisebox{-0.5\height}{\includegraphics[width=0.25\linewidth]{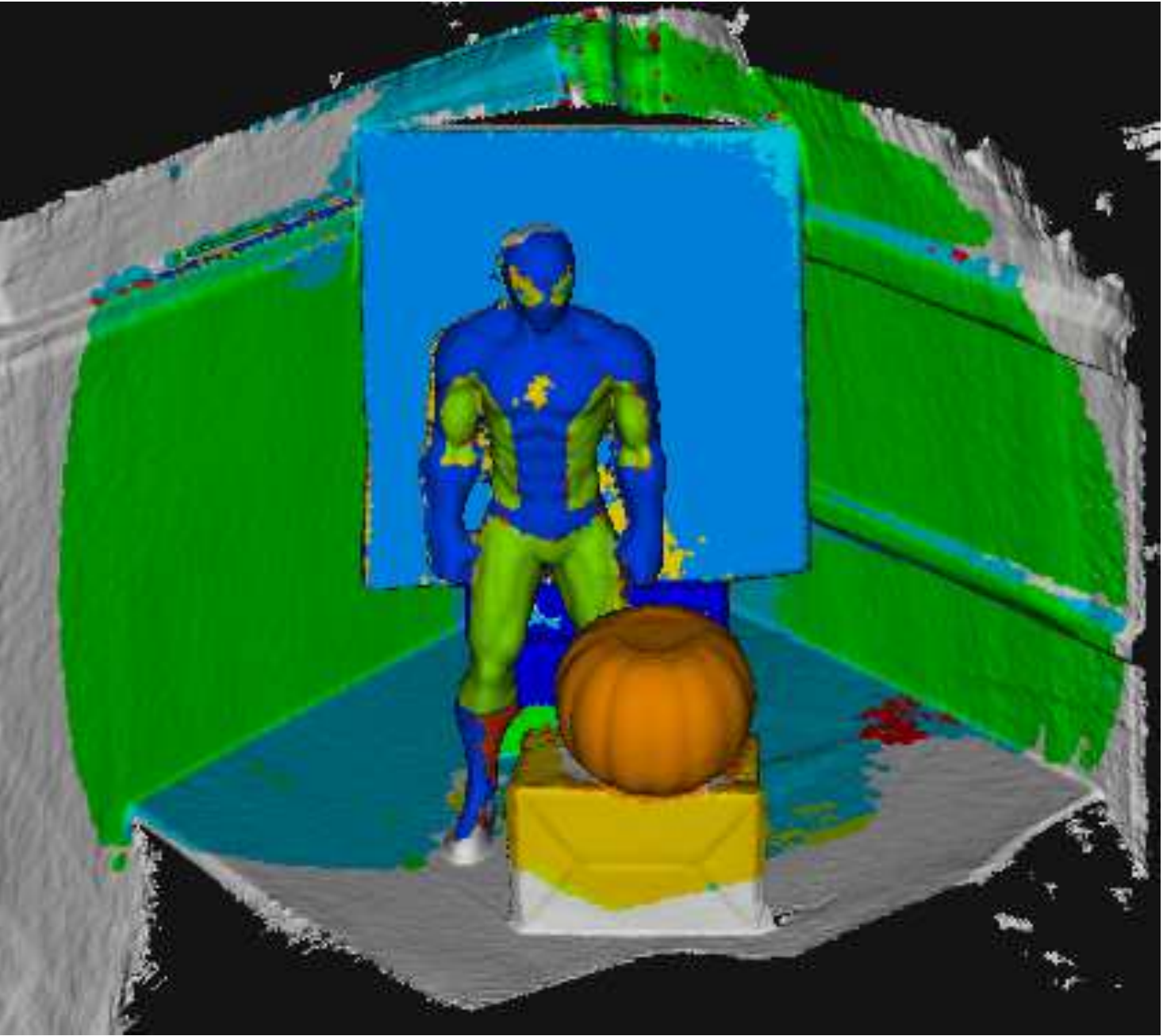}} & \raisebox{-0.5\height}{\includegraphics[width=0.25\linewidth]{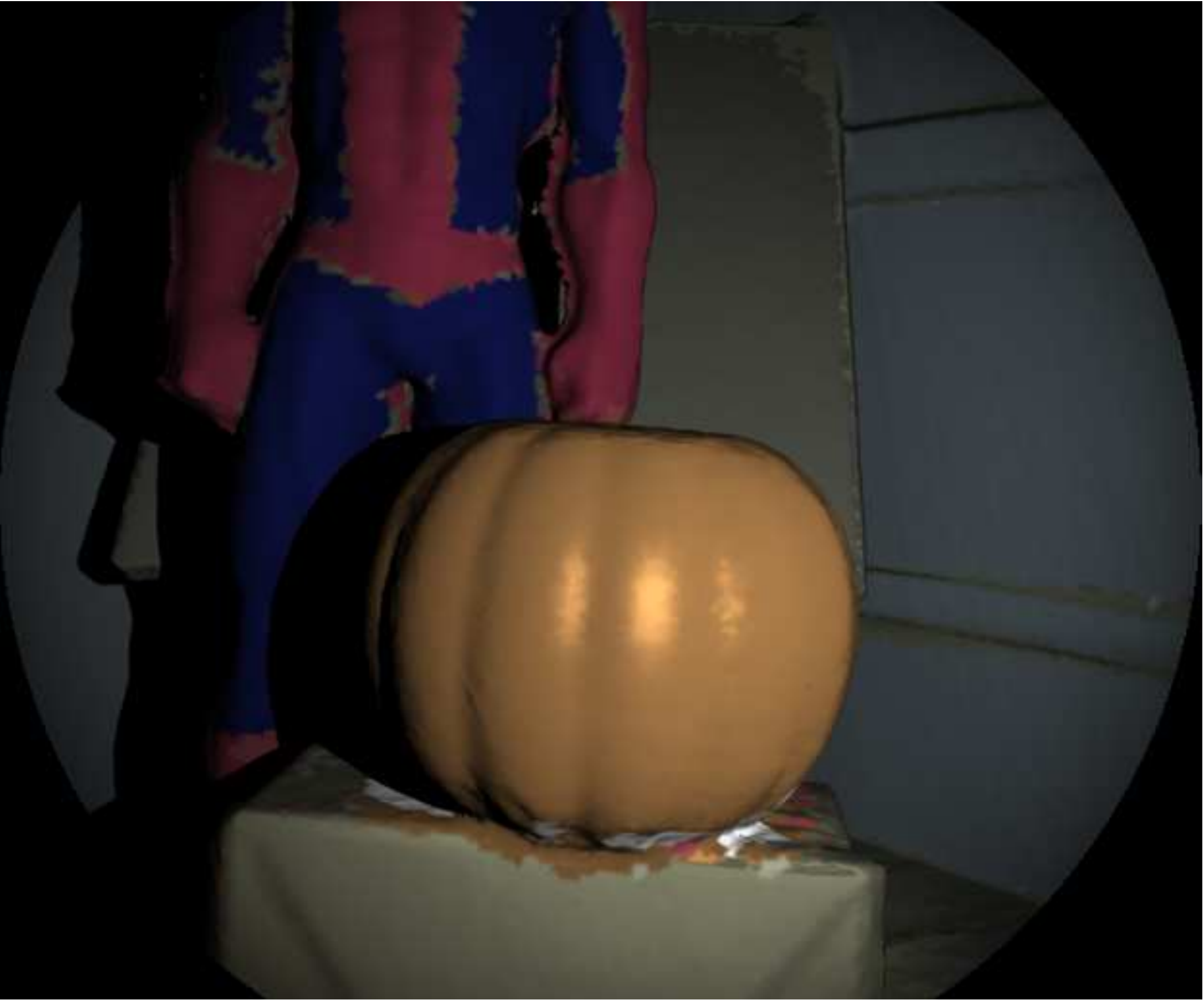}} & \raisebox{-0.5\height}{\includegraphics[width=0.25\linewidth]{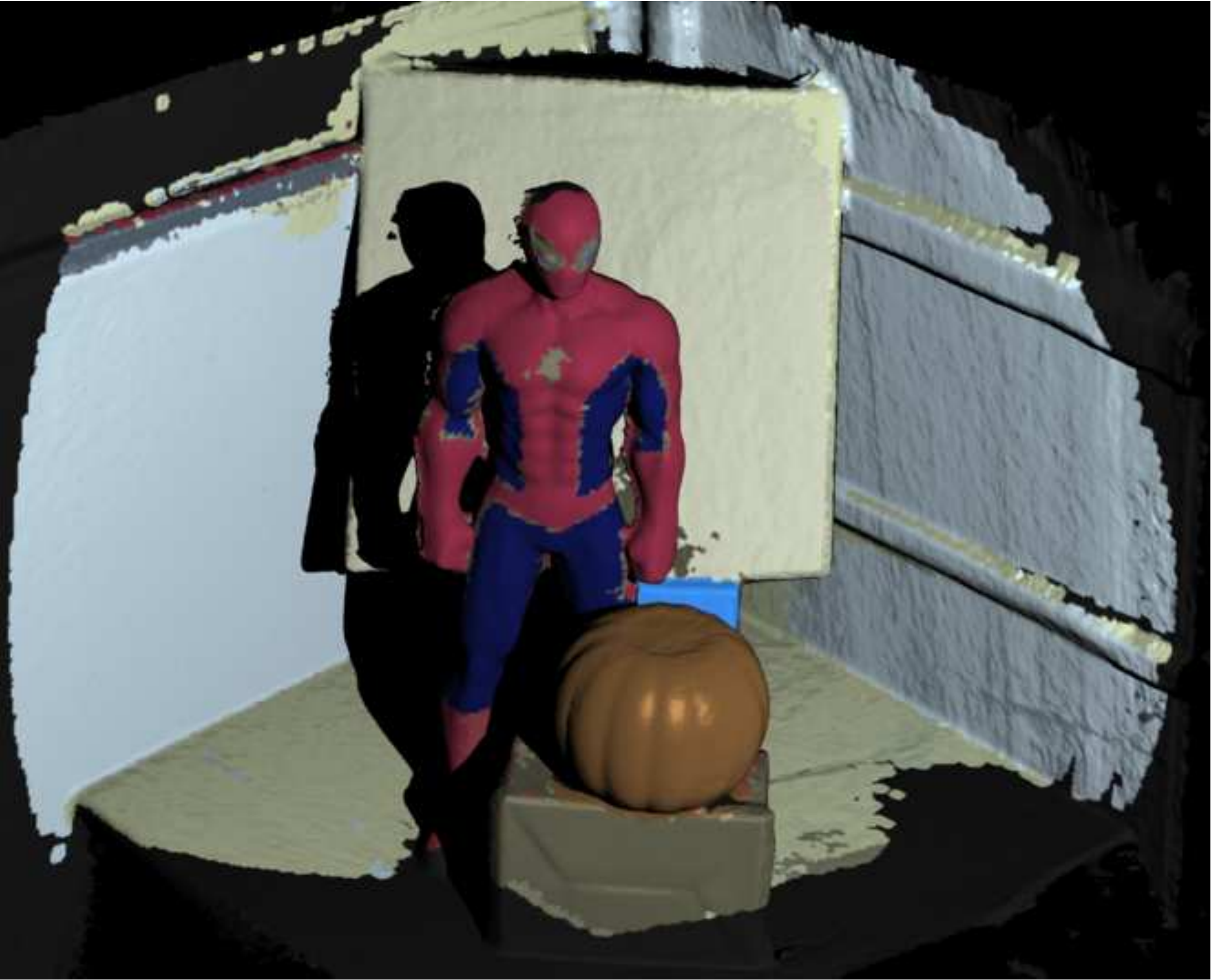}} \\
(b) & (d) & (f) & (h)
\end{tabular}
\addtolength{\tabcolsep}{4pt}  
\end{center}
\caption{This figure shows results of the scene \textit{Spiderman} in a similar way as \figref{gymball_results} and \figref{pot_results}. The 8 materials in (g) correspond to pumpkin, red region on spiderman, blue region on spiderman, board behind spiderman, box behind the spiderman's legs, box in front of spiderman, wall and the supporting board under spiderman.}
\label{spiderman_results}
\vspace{-5mm}
\end{figure*}

\figref{spiderman_results} shows results of a scene \textit{Spiderman}. Our system still successfully segmented vertices of different materials, as is shown in (c-d). Out of all 13 material groups produced by our system, there are 8 dominant groups having considerably more vertices than the rest. Rendering of the 8 materials is shown in (g). Notice that for some rendering, there is a circular artifact. This is because for that material, our system acquires only a portion of its BRDF and we use a naive interpolation/extrapolation method to complete the BRDF. Obviously, The simple interpolation/extrapolation strategy does not guarantee smooth rendering.

\noindent
\textbf{Limitations} As can be seen in \figref{pot_results}(e-f), the highlight of the brown pot in the original IR image was smoothed out in the re-rendering. This is because sharp specularity can only be faithfully captured with highly accurate calibration of surface normal, lighting and viewing directions, which are impossible for our system. Due to the same reason, our system cannot handle surfaces with much texture well.

\section{Conclusion and Future Work}
In this paper, we present a novel device for capturing both shape and BRDFs.
Unlike traditional appearance capture system, our device features not only portability and ease of use, but also the capability of acquiring the appearance of a scene instead of a single object.

The major difficulty in building such a system lies in handling the incomplete and severely corrupted BRDF observations, which was seldom seen in previous systems thanks to highly controlled environments.
To solve the problem, we propose the material segmentation propagation algorithm, which
automatically segments vertices into different material groups by using those sparse and noisy BRDF samples.
We see our device as a stepping stone to more capable RGBD-M(aterial) scanners.

Currently our system processes data offline. We hope to make it work and give feedback in real-time, which is important in guiding the user for better scanning.

{\small
\bibliographystyle{ieee}
\bibliography{egbib}

\begin{thebibliography}{10}\itemsep=-1pt

\bibitem{aittala2013practical}
M.~Aittala, T.~Weyrich, and J.~Lehtinen.
\newblock Practical svbrdf capture in the frequency domain.
\newblock {\em ACM Trans. Graph.}, 32(4):110, 2013.

\bibitem{alldrin2008photometric}
N.~Alldrin, T.~Zickler, and D.~Kriegman.
\newblock Photometric stereo with non-parametric and spatially-varying
  reflectance.
\newblock In {\em {Proc. CVPR}}, pages 1--8. IEEE, 2008.

\bibitem{CPTK14cvpr}
G.~Choe, J.~Park, Y.-W. Tai, and I.~Kweon.
\newblock Exploiting shading cues in kinect ir images for geometry refinement.
\newblock In {\em {Proc. CVPR}}, 2014.

\bibitem{dam1998quaternions}
E.~B. Dam, M.~Koch, and M.~Lillholm.
\newblock {\em Quaternions, interpolation and animation}.
\newblock Datalogisk Institut, K{\o}benhavns Universitet, 1998.

\bibitem{furukawa2010accurate}
Y.~Furukawa and J.~Ponce.
\newblock Accurate, dense, and robust multiview stereopsis.
\newblock {\em {IEEE Trans. Pattern Anal. Mach. Intell.}}, 32(8):1362--1376,
  2010.

\bibitem{goldman2010shape}
D.~B. Goldman, B.~Curless, A.~Hertzmann, and S.~M. Seitz.
\newblock Shape and spatially-varying brdfs from photometric stereo.
\newblock {\em {IEEE Trans. Pattern Anal. Mach. Intell.}}, 32(6):1060--1071,
  2010.

\bibitem{haque2014high}
S.~Haque, A.~Chatterjee, V.~M. Govindu, et~al.
\newblock High quality photometric reconstruction using a depth camera.
\newblock In {\em {Proc. CVPR}}, pages 2283--2290. IEEE, 2014.

\bibitem{hernandez2008multiview}
C.~Hern{\'a}ndez, G.~Vogiatzis, and R.~Cipolla.
\newblock Multiview photometric stereo.
\newblock {\em {IEEE Trans. Pattern Anal. Mach. Intell.}}, 30(3):548--554,
  2008.

\bibitem{holroyd2010coaxial}
M.~Holroyd, J.~Lawrence, and T.~Zickler.
\newblock A coaxial optical scanner for synchronous acquisition of 3d geometry
  and surface reflectance.
\newblock {\em {ACM Trans. Graph. (Proc. of SIGGRAPH)}}, 29(4):99, 2010.

\bibitem{izadi2011kinectfusion}
S.~Izadi, D.~Kim, O.~Hilliges, D.~Molyneaux, R.~Newcombe, P.~Kohli, J.~Shotton,
  S.~Hodges, D.~Freeman, A.~Davison, et~al.
\newblock Kinectfusion: real-time 3d reconstruction and interaction using a
  moving depth camera.
\newblock In {\em Proceedings of the 24th annual ACM symposium on User
  interface software and technology}, pages 559--568. ACM, 2011.

\bibitem{lensch2003image}
H.~Lensch, J.~Kautz, M.~Goesele, W.~Heidrich, and H.-P. Seidel.
\newblock Image-based reconstruction of spatial appearance and geometric
  detail.
\newblock {\em {ACM Trans. Graph. (Proc. of SIGGRAPH)}}, 22(2):234--257, 2003.

\bibitem{newcombe2011dtam}
R.~A. Newcombe, S.~J. Lovegrove, and A.~J. Davison.
\newblock Dtam: Dense tracking and mapping in real-time.
\newblock In {\em {Proc. ICCV}}, pages 2320--2327. IEEE, 2011.

\bibitem{noll2014robust}
T.~N{\"o}ll, J.~K{\"o}hler, and D.~Stricker.
\newblock Robust and accurate non-parametric estimation of reflectance using
  basis decomposition and correction functions.
\newblock In {\em {Proc. ECCV}}, pages 376--391. Springer, 2014.

\bibitem{romeiro2008passive}
F.~Romeiro, Y.~Vasilyev, and T.~Zickler.
\newblock Passive reflectometry.
\newblock In {\em {Proc. ECCV}}, pages 859--872. Springer, 2008.

\bibitem{rusinkiewicz1998new}
S.~M. Rusinkiewicz.
\newblock A new change of variables for efficient brdf representation.
\newblock In {\em Rendering techniques’ 98}, pages 11--22. Springer, 1998.

\bibitem{sato1997object}
Y.~Sato, M.~D. Wheeler, and K.~Ikeuchi.
\newblock Object shape and reflectance modeling from observation.
\newblock In {\em Proceedings of the 24th annual conference on Computer
  graphics and interactive techniques}, pages 379--387. ACM
  Press/Addison-Wesley Publishing Co., 1997.

\bibitem{tunwattanapong2013acquiring}
B.~Tunwattanapong, G.~Fyffe, P.~Graham, J.~Busch, X.~Yu, A.~Ghosh, and
  P.~Debevec.
\newblock Acquiring reflectance and shape from continuous spherical harmonic
  illumination.
\newblock {\em {ACM Trans. Graph. (Proc. of SIGGRAPH)}}, 32(4):109, 2013.

\bibitem{wang2008modeling}
J.~Wang, S.~Zhao, X.~Tong, J.~Snyder, and B.~Guo.
\newblock Modeling anisotropic surface reflectance with example-based
  microfacet synthesis.
\newblock In {\em ACM Transactions on Graphics (TOG)}, volume~27, page~41. ACM,
  2008.

\bibitem{wu2014real}
C.~Wu, M.~Zollh{\"o}fer, M.~Nie{\ss}ner, M.~Stamminger, S.~Izadi, and
  C.~Theobalt.
\newblock Real-time shading-based refinement for consumer depth cameras.
\newblock {\em Proc. SIGGRAPH Asia}, 2014.

\bibitem{wu2015appfusion}
H.~Wu and K.~Zhou.
\newblock Appfusion: Interactive appearance acquisition using a kinect sensor.
\newblock In {\em Computer Graphics Forum}. Wiley Online Library, 2015.

\bibitem{wu2013calibrating}
Z.~Wu and P.~Tan.
\newblock Calibrating photometric stereo by holistic reflectance symmetry
  analysis.
\newblock In {\em {Proc. CVPR}}, pages 1498--1505. IEEE, 2013.

\bibitem{yu2013shading}
L.-F. Yu, S.-K. Yeung, Y.-W. Tai, and S.~Lin.
\newblock Shading-based shape refinement of rgb-d images.
\newblock In {\em {Proc. CVPR}}, pages 1415--1422. IEEE, 2013.

\bibitem{yu1999inverse}
Y.~Yu, P.~Debevec, J.~Malik, and T.~Hawkins.
\newblock Inverse global illumination: Recovering reflectance models of real
  scenes from photographs.
\newblock In {\em Proceedings of the 26th annual conference on Computer
  graphics and interactive techniques}, pages 215--224. ACM
  Press/Addison-Wesley Publishing Co., 1999.

\bibitem{zhou2013dense}
Q.-Y. Zhou and V.~Koltun.
\newblock Dense scene reconstruction with points of interest.
\newblock {\em {ACM Trans. Graph. (Proc. of SIGGRAPH)}}, 32(4):112, 2013.

\bibitem{zhou2014color}
Q.-Y. Zhou and V.~Koltun.
\newblock Color map optimization for 3d reconstruction with consumer depth
  cameras.
\newblock {\em {ACM Trans. Graph. (Proc. of SIGGRAPH)}}, 33(4):155, 2014.

\bibitem{zhou2013multi}
Z.~Zhou, Z.~Wu, and P.~Tan.
\newblock Multi-view photometric stereo with spatially varying isotropic
  materials.
\newblock In {\em {Proc. CVPR}}, pages 1482--1489. IEEE, 2013.

\bibitem{zollhofershading}
M.~Zollh{\"o}fer, A.~Dai, M.~Innmann, C.~Wu, M.~Stamminger, C.~Theobalt, and
  M.~Nie{\ss}ner.
\newblock Shading-based refinement on volumetric signed distance functions.
\newblock {\em Proc. SIGGRAPH}, 2015.

\end{thebibliography}
}

\end{document}